\documentclass[runningheads]{llncs}

\usepackage{eccv}

\usepackage{eccvabbrv}

\usepackage{graphicx}
\usepackage{booktabs}
\usepackage{tabularx}
\usepackage{array}
\usepackage{caption}
\usepackage{makecell}
\usepackage{multirow}
\usepackage[subtle,paragraphs=normal,mathdisplays=normal,wordspacing=normal]{savetrees}
\usepackage[accsupp]{axessibility}  
\usepackage[table]{xcolor}   
\definecolor{wipblue}{RGB}{222,235,247}  

\definecolor{best}{HTML}{F6AD55}   
\definecolor{second}{HTML}{FBD38D} 
\usepackage{xcolor}
\definecolor{ForestGreenCustom}{HTML}{228B22}
\newcommand{\highlight}[1]{\textcolor{ForestGreenCustom}{#1}}
\usepackage{xcolor}

\usepackage{hyperref}
\usepackage{wrapfig}
\usepackage{cuted}

\usepackage{orcidlink}

\usepackage{enumitem}
\setlist{nosep}
\setlist{topsep=0pt, itemsep=0pt, partopsep=0pt, parsep=0pt}

\begin{document}

\title{FlexiAvatar: Unified 3D Gaussian Human Avatars Under Arbitrary Body Visibility} 
\titlerunning{FlexiAvatar: Unified 3D Gaussian Human Avatars}

\author{Yihalem~Yimolal~Tiruneh\inst{1} \and
Muhammad~Salman~Ali\inst{1} \and
Uyoung~Jeong\inst{1} \and
Muneeb~A.~Khan\inst{1} \and
MD~Khalequzzaman~Chowdhury~Sayem\inst{1} \and
Allanur~Bayramgeldiyev\inst{1} \and
Binod~Bhattarai\inst{2,3,4} \and
Seungryul~Baek\inst{1}}

\authorrunning{Y.~Tiruneh et al.}

\institute{%
{\small
$^{1}$UNIST, South Korea \enspace
$^{2}$University of Aberdeen, UK \enspace
$^{3}$University College London, UK \enspace
$^{4}$Fogsphere, UK}
}

\maketitle
\vspace{-0.3cm}
\begin{center}
  \vspace{-0.7em}
  \includegraphics[width=\textwidth]{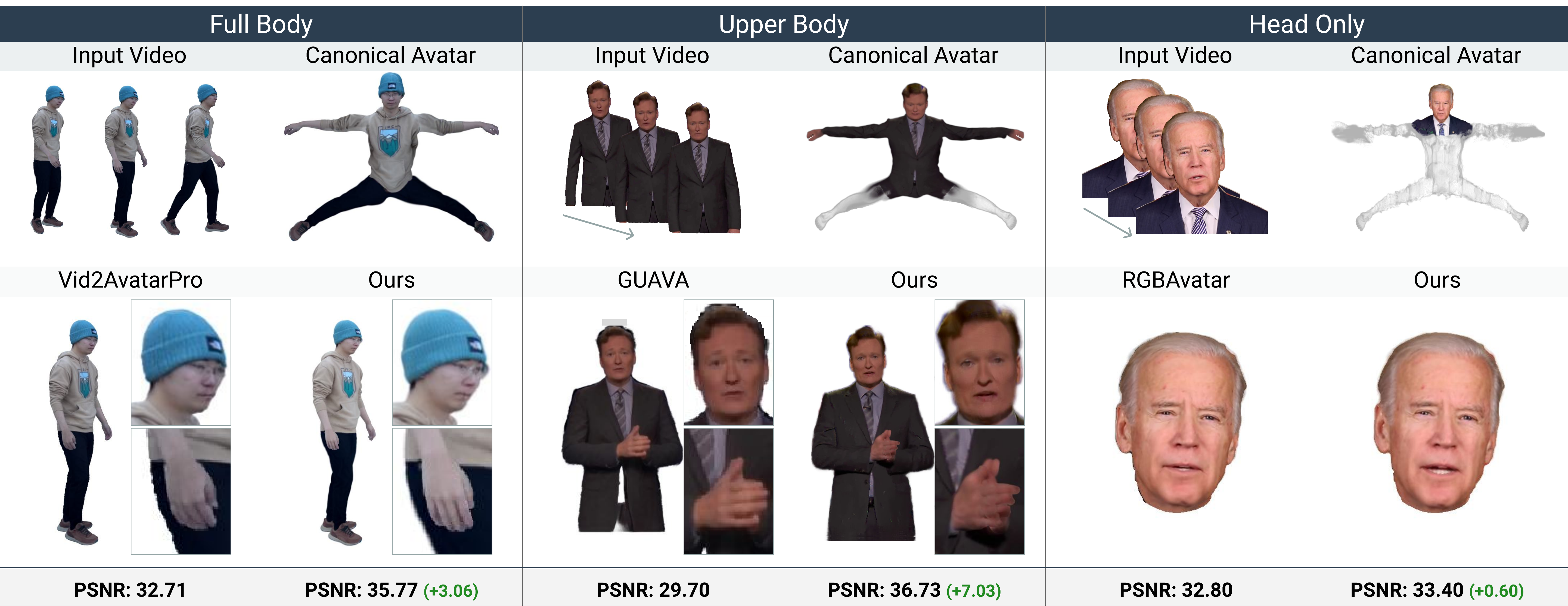}
  \vspace{-1.3em}
  \captionof{figure}{From monocular video under arbitrary body visibility, FlexiAvatar reconstructs animatable 3D Gaussian avatars within a single pipeline, outperforming state-of-the-art full-body, upper-body, and head-only methods.}
  \label{fig:main_teaser}
  \vspace{-1.0em}
\end{center}
\begin{abstract}
Reconstructing animatable 3D human avatars from monocular video is a fundamental problem in computer vision with broad applications in AR/VR and digital content creation. Existing approaches typically couple parametric body models with neural rendering or 3D Gaussian splatting and optimize all body regions jointly from short videos, which often degrades fidelity in the visible areas. To overcome this limitation, we introduce \textbf{FlexiAvatar}, a unified framework that explicitly optimizes only the visible body regions, effectively eliminating artifacts arising from unobserved limbs. Our method integrates occlusion-robust SMPL-X tracking with part-specific residual refinement to capture high-frequency geometric and appearance details. To complete entirely unseen regions (e.g., back views), we leverage a diffusion-based approach to generate texture consistent with the observed appearance. Experiments on full-body (NeuMan, ZJU-MoCap, WildAvatar), upper/half-body (talk-show clips), and head-only (INSTA) inputs show that FlexiAvatar delivers consistently higher reconstruction quality, outperforming state-of-the-art methods by an average PSNR improvement of approximately 3\% across datasets. Finally, by restricting optimization to observed regions, our method reduces the effective number of Gaussians that must be optimized and rendered, leading to reduced runtime and memory overhead in partial-visibility scenarios. The project page can be found \href{https://yihalem1.github.io/FlexiAvatar/}{here}.
\end{abstract}
    
\section{Introduction}

 Realistic, animatable human avatars are essential for applications such as AR/VR~\cite{shen2023xavatar, ma2021pixel, INSTA:CVPR23}, film production, remote collaboration~\cite{piumsomboon2018mini, fidalgo2023manipulating}, and interactive entertainment~\cite{kang2023super}. However, most real-world and in-the-wild video data (video calls, talk-show clips, and social media footage) spans a wide visibility spectrum, containing severe cropping, partial-body visibility, and occlusions, rather than clean, full-body studio recordings. Reconstructing a complete, high-fidelity, animatable avatar from such monocular partial-body input remains an open and largely unsolved challenge.

Classical approaches such as SMPL~\cite{Loper:SMPL:TOG15}, FLAME~\cite{Li:FLAME:SA17}, and SMPL-X~\cite{Pavlakos:SMPLX:CVPR19} employ parametric mesh models that provide strong priors for articulated body, face, and hand reconstruction. However, these models are inherently constrained by their fixed topology and often struggle to reproduce fine-grained geometric and textural fidelity~\cite{AlldieckBodyNet:3DV19, Zheng2023FLAME}. Neural rendering methods such as NeRF~\cite{Mildenhall:NeRF:ECCV20} improve visual realism over parametric approaches, but typically require lengthy per-subject optimization and lack real-time rendering efficiency~\cite{Kopanas:PointBased:CVPR21}.

3D Gaussian Splatting (3DGS)~\cite{Kerbl:3DGS:SIGA23} has enabled fast, high-quality reconstruction of dynamic scenes~\cite{ali2026compression}. Building on this representation, avatar methods such as GART~\cite{Lei:GART:CVPR24}, GaussianAvatar~\cite{Hu:GaussianAvatar:CVPR24},  ExAvatar~\cite{Moon:ExAvatar:ECCV24} and Vid2AvatarPro~\cite{Guo:Vid2AvatarPro:CVPR25} achieve high visual fidelity and strong animation quality. Despite these advances, all existing 3DGS avatar methods share a common assumption: the entire body is visible throughout training. 
In portrait recordings, video calls, and talk-show clips where only the face, head, or upper body ever appears, this assumption is systematically violated. Continuing to supervise Gaussians for non-visible limbs introduces hallucinated geometry and texture drift that corrupts even the visible regions.  No existing method ~\cite{Moon:ExAvatar:ECCV24, Guo:Vid2AvatarPro:CVPR25} directly targets this visibility spectrum problem within the optimization loop itself.

Portrait-level approaches such as InsTaG~\cite{InsTaG:CVPR25}, HRAvatar~\cite{Zhang:HRAvatar:CVPR25}, RGBAvatar~\cite{RGBAvatar:CVPR25}, and FATE~\cite{Zhang:FATE:CVPR25} achieve high-fidelity reconstruction of the head region. In contrast, upper-body methods such as GUAVA~\cite{Zhang:GUAVA:ICCV25} extend the reconstruction scope beyond the face; however, they remain constrained to a fixed upper-body input setting and often exhibit degraded quality in occluded or unseen regions. Neither class of methods generalizes across the entire visibility spectrum: full-body methods~\cite{Moon:ExAvatar:ECCV24, Lei:GART:CVPR24, Guo:Vid2AvatarPro:CVPR25, Xie:Wild2Avatar:CVPR23} optimize all body regions regardless of visibility, inevitably hallucinating geometry and appearance in unobserved areas, while localized methods, whether head-only~\cite{InsTaG:CVPR25, Zhang:FATE:CVPR25, Qian:GaussianAvatars:CVPR24} or upper-body~\cite{Zhang:GUAVA:ICCV25} are architecturally constrained to a specific body coverage input setting. This fragmentation forces practitioners to maintain separate, purpose-built pipelines for each input visibility setting, a significant practical barrier to real-world deployment~\cite{cho2021dynamic}.

To address these limitations, we propose FlexiAvatar, a visibility-aware Gaussian avatar reconstruction framework that explicitly restricts optimization to observed body regions, updating geometry and appearance only where visual evidence exists. This design eliminates ghost geometry and texture drift, and enables high-quality animation from full-body, upper-body, and head-only inputs within a single unified pipeline, as illustrated in Fig.~\ref{fig:main_teaser}. Our method combines structured priors from SMPL-X with region-level part-specific residual refinement and a visibility-guided learning objective. We employ occlusion-robust SMPL-X tracking to maintain accurate articulation across varying visibility conditions, and introduce part-specific residual modules to recover high-frequency appearance details, such as facial expressions and finger creases, that exceed the capacity of global Gaussian representations. For unobserved regions such as the back, we leverage a diffusion-based approach to generate subject-consistent auxiliary views, providing appearance supervision for unseen areas without introducing hallucinations or propagating errors to the observed regions. By integrating (i) visibility-aware optimization, (ii) occlusion-robust SMPL-X tracking, (iii) part-specific residual refinement, and (iv) generative texture completion, FlexiAvatar achieves high-quality reconstruction across varying visibility (full-body, upper-body, and head-only) inputs within a unified pipeline.

\begin{figure*}[t!]
\centering
\includegraphics[width=\linewidth]{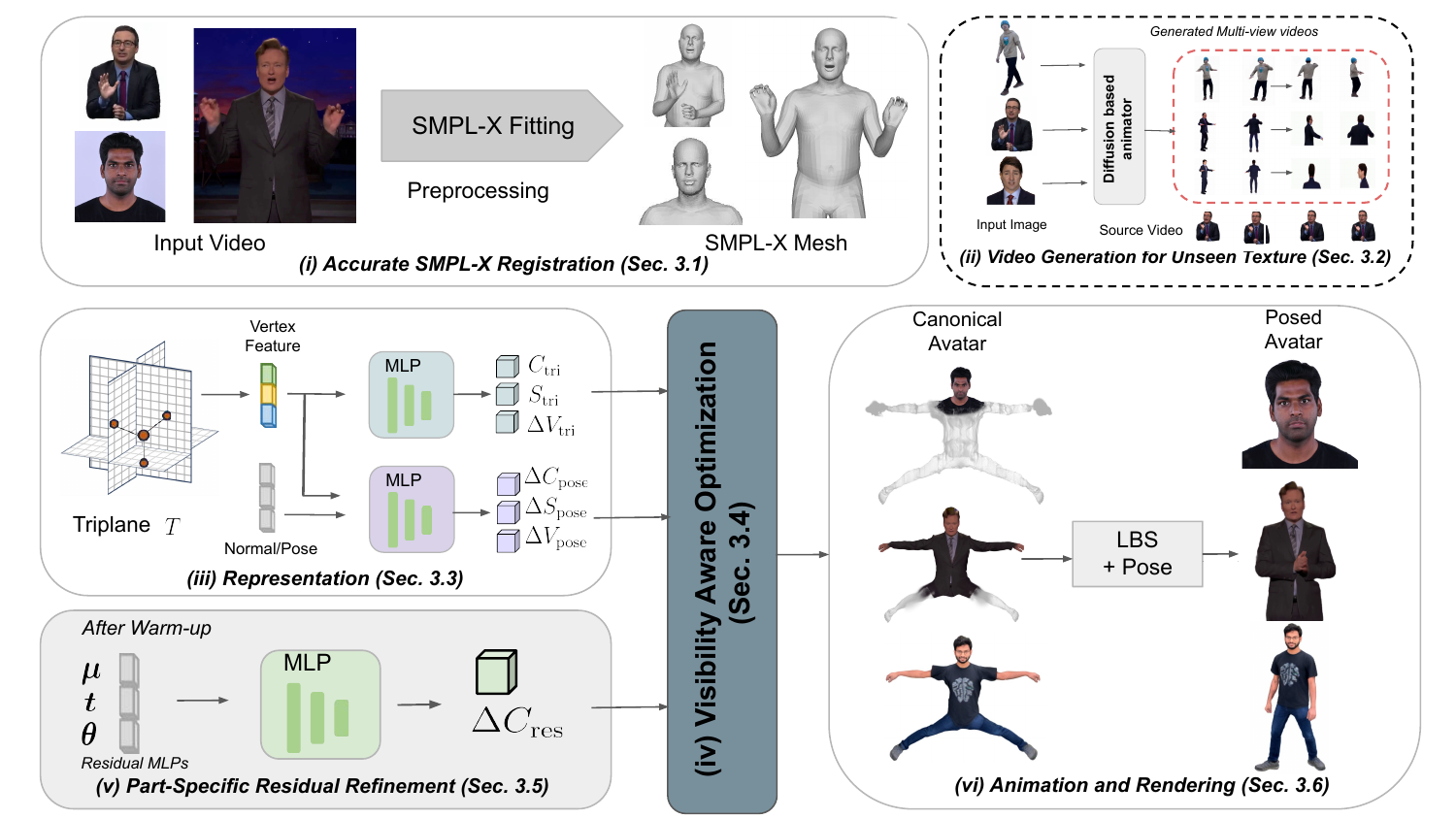}
\caption{Overview of our \textit{FlexiAvatar} pipeline. From a monocular video, we perform visibility-aware SMPL-X registration (\cref{subsec:smplx_registration}) and generate auxiliary multi-view sequences to supervise unseen regions (\cref{subsec:video_generation}). Canonical vertices are projected onto a learnable triplane whose MLPs regress static and pose-dependent Gaussian parameters (\cref{subsec:representation}). After warm-up, visibility-aware optimization prunes low-evidence Gaussians (\cref{subsec:vis_aware}) and residual MLPs enhance detail (\cref{subsec:refinement}). Gaussians are animated by SMPL-X linear blend skinning and rendered with 3DGS (\cref{subsec:animcation_rendering}).}    
    \vspace{-0.3cm}
\label{fig:overview_pipeline_figure}
\end{figure*}

Our main contributions are:
\begin{itemize}
\item We propose FlexiAvatar, a visibility-aware Gaussian avatar reconstruction framework that restricts optimization to visually observed body regions, updating geometry and appearance only where image evidence is present. This formulation eliminates hallucinated artifacts and texture drift, and enables high-quality animation from full-body, upper-body, or head-only inputs (Sec.~\ref{sec:methodology}).
\item Our method incorporates an occlusion-robust SMPL-X tracking pipeline for accurate body registration, residual modules to recover high-frequency facial and hand details, and a diffusion-based generative texture completion module that synthesizes novel views to fill unseen regions (Fig.~\ref{fig:overview_pipeline_figure}).
\item Extensive quantitative and qualitative experiments across full-body~\cite{Jiang:NeuMan:ECCV22,zjumocap_dataset,wildavatar:CVPR2025}, upper-body~\cite{Yi:TalkShow:CVPR23}, and head-only~\cite{INSTA:CVPR23} datasets demonstrate that FlexiAvatar consistently outperforms state-of-the-art methods (Sec.~\ref{sec:experiments}).
\end{itemize} 

\vspace{-2mm}
\section{Related Work}
\noindent\textbf{3D-based Avatar Reconstruction.}
\label{sec:rw_3d}
Template-based parametric models like FLAME~\cite{Li:FLAME:SA17}, SMPL~\cite{Loper:SMPL:TOG15},  and SMPL-X~\cite{Pavlakos:SMPLX:CVPR19} provide controllable body, face, and hand representations and remain standard backbones in monocular human-capture pipelines. Recent advances in neural rendering have shifted from implicit NeRF-style fields to explicit representations to improve fidelity and enable differentiable primitives. 3DGS~\cite{Kerbl:3DGS:SIGA23} further enables real-time, high-fidelity rendering through an explicit representation, making it a strong foundation for animatable avatars.

Person-specific Gaussian avatars from monocular video achieve high fidelity and motion controllability, yet uniformly assume full-body visibility during training. Early methods like Vid2Avatar~\cite{Guo:Vid2Avatar:CVPR23} introduced self-supervised scene decomposition to reconstruct avatars from unmasked in-the-wild videos. 3DGS-Avatar~\cite{Qian:3DGSAvatar:CVPR24} leverages a non-rigid deformation network with 3D Gaussian Splatting to achieve fast training and real-time rendering of clothed avatars from monocular video. GART~\cite{Lei:GART:CVPR24} employs template-driven Gaussians with learnable skinning for articulation, while 
GaussianAvatar~\cite{Hu:GaussianAvatar:CVPR24} introduces pose-conditioned deformation for dynamic appearance. ExAvatar~\cite{Moon:ExAvatar:ECCV24} integrates Gaussian splats to the SMPL-X topology for improved facial and hand expressivity. GauHuman~\cite{gauhuman:CVPR2024} employs body-prior-guided Gaussian splatting for efficient high-fidelity avatar reconstruction from monocular video, while ToMiE~\cite{tomie:ICCV2025} decomposes texture and geometry into a modular implicit representation, and Vid2AvatarPro\cite{Guo:Vid2AvatarPro:CVPR25} incorporates universal generative priors for in-the-wild texture completion. Under partial views, all these methods hallucinate unseen regions, introduce artifacts, or propagate errors to visible areas.

For portrait-level avatar reconstruction, GaussianAvatars~\cite{Qian:GaussianAvatars:CVPR24} employs rigged 3DGS; InsTaG~\cite{InsTaG:CVPR25} learns personalized talking heads from short monocular sequences; HRAvatar~\cite{Zhang:HRAvatar:CVPR25} yields relightable head avatars; and FATE~\cite{Zhang:FATE:CVPR25} supports $360^\circ$ full-head reconstruction with texture editing. RGBAvatar~\cite{RGBAvatar:CVPR25} compresses Gaussian blendshapes for online modeling, GEM~\cite{Zielonka:GEM:CVPR25} uses lightweight eigen-bases, and  MeGA~\cite{Wang:MeGA:CVPR25} adopts component-wise modeling. While these approaches achieve high fidelity in head avatars, they remain limited to localized reconstruction and do not generalize to full or upper-body avatars.

Recently, generalizable upper or half-body Gaussian avatars from \emph{minimal} input have gained attention. GUAVA~\cite{Zhang:GUAVA:ICCV25} reconstructs an animatable upper-body avatar from a \emph{single} image via inverse texture mapping and neural refinement, while AniGS~\cite{Qiu:AniGS:CVPR25} uses inconsistent-Gaussian reconstruction to improve generalization. However, single-image pipelines lack multi-view constraints and often struggle with unseen-region completion and visual fidelity. Our approach bridges this gap by preserving the practicality of sparse inputs while introducing visibility-aware optimization that focuses on observed geometry, mitigating hallucination of unobserved regions and enabling stable optimization and reconstruction under partial views.

\noindent\textbf{2D-based Human Animation.}
\label{sec:rw_2d}
2D motion transfer and diffusion-based generation methods provide strong appearance priors but lack a persistent 3D identity representation. DreamPose~\cite{Karras:DreamPose:ICCV23} synthesizes fashion videos from a single image and pose, while MagicPose~\cite{MagicPose} enables identity-aware pose and expression retargeting through latent diffusion. Champ~\cite{Zhu:Champ:ECCV24} introduces SMPL-conditioned depth and normal priors for controllable animation, and MimicMotion~\cite{Zhang:MimicMotion:ICML25} extends this direction with confidence-aware pose guidance for smooth long-term motion. Unlike these image-space methods, we leverage a diffusion-based approach to generate texture-consistent auxiliary views for unobserved regions.

\vspace{-2mm}
\section{Methodology}
\label{sec:methodology}
Our proposed method, FlexiAvatar, reconstructs a high-fidelity, animatable 3D avatar from a monocular video, even under partial-body visibility (e.g., upper-body or head-only). Our method consists of six components (as illustrated in ~\cref{fig:overview_pipeline_figure}):
(i) performing accurate, visibility-aware SMPL-X registration to obtain a personalized template (\cref{subsec:smplx_registration});
(ii) augmenting the input with diffusion-generated views to complete unseen textures (\cref{subsec:video_generation});
(iii) learning a triplane-conditioned hybrid mesh–Gaussian representation for geometry and appearance (\cref{subsec:representation});
(iv) enforcing visibility-aware optimization to prevent hallucination (\cref{subsec:vis_aware});
(v) refining high-frequency facial and hand appearance with a part-specific residual module (\cref{subsec:refinement});  and
(vi) animating the avatar with SMPL-X poses and rendering it using 3D Gaussian Splatting (\cref{subsec:animcation_rendering}).

\vspace{-2mm}
\subsection{Accurate SMPL-X Registration}
\label{subsec:smplx_registration}
Given an input sequence, we extract frame-wise SMPL-X~\cite{Pavlakos:SMPLX:CVPR19} parameters and 2D whole-body keypoints using off-the-shelf estimators~\cite{smplest, dwpose}. These provide an initialization of body pose $\mathbf{\theta} \in \mathbb{R}^{55\times3}$, shape $\mathbf{\beta} \in \mathbb{R}^{100}$, and facial expression $\mathbf{\psi} \in \mathbb{R}^{50}$, along with keypoints $\mathbf{K} \in \mathbb{R}^{J\times2}$ and confidence scores $\mathbf{s} \in \mathbb{R}^{J}$.
Directly using all detected keypoints leads to artifacts when large portions of the body are outside the camera’s field of view. To mitigate this, we introduce a binary visibility mask that excludes low-confidence or occluded joints from the reprojection objective. This ensures the stable fitting of the observed body parts without distorting unobserved regions.
To capture identity-specific geometric nuances, we further optimize joint offsets $\Delta \mathbf{J}$ and facial offsets $\Delta \mathbf{V}_{\text{face}}$, producing a personalized canonical template.  The final objective is:
\vspace{-0.15cm}
\begin{equation}
\mathcal{L}_{\text{regi}} = \mathcal{L}_{\text{kpt}} + \lambda_{\text{init}} \mathcal{L}_{\text{init}} + \lambda_{\text{sface}} \mathcal{L}_{\text{sface}} + \lambda_{\text{sreg}} \mathcal{L}_{\text{sreg}},
\label{eq:tracking_loss}
\end{equation}

\noindent
where $\mathcal{L}_{\text{regi}}$ denotes the overall SMPL-X registration objective, $\mathcal{L}_{\text{kpt}}$ is the 2D keypoint reprojection loss, $\mathcal{L}_{\text{init}}$ penalizes deviations from the off-the-shelf initialization, $\mathcal{L}_{\text{sface}}$ aligns the SMPL-X face geometry to a pre-fitted FLAME template to preserve identity-specific facial details and $\mathcal{L}_{\text{sreg}}$ regularizes both the shape symmetry and the learned joint-offset parameters. To improve robustness under occlusion, we apply a masking strategy to the 2D reprojection loss $\mathcal{L}_{\text{kpt}}$, defined as:

\vspace{-0.2cm}
\begin{equation}
\mathcal{L}_{\text{kpt}} = \sum_{j} \mathbf{m}_j \left\| \Pi\!\big(V_j(\mathbf{\theta}, \mathbf{\beta}, \Delta J)\big) - \mathbf{K}_j \right\|_1,
\label{eq:Keypoint_loss}
\end{equation}
\vspace{-0.2cm}

\noindent
where $\Pi$ is the camera projection function, $\mathbf{V}_j$ is the $j$-th 3D joint location derived from the personalized SMPL-X model (which includes parameters $\theta$, $\beta$, and the joint offset $\Delta \mathbf{J}$), $\mathbf{K}_j$ is the corresponding target 2D keypoint, and $\mathbf{m}_j \in \{0,1\}$ is a per-joint binary mask 
derived from detector confidence scores. Here $\mathbf{m} \in \mathbb{R}^{J}$ zeros out low-confidence or 
occluded joints that fall below the threshold $\tau$, restricting supervision to well-localized, visible keypoints only and preventing unreliable 2D evidence from corrupting the pose update. Based on empirical evidence, we set $\tau = 0.4$. $\mathcal{L}_{\text{init}}$ is defined as:
\begin{equation}
\mathcal{L}_{\text{init}} = \left\| \mathbf{\theta} - \mathbf{\theta}^* \right\|_1 + \left\| \mathbf{\beta} - \mathbf{\beta}^* \right\|_1 + \left\| \mathbf{\psi} - \mathbf{\psi}^* \right\|_1,
\end{equation}

\noindent
$\mathcal{L}_{\text{init}}$ term acts as a strong prior, leveraging the robust output of the initial regressor and preventing the optimization from drifting into implausible states. $\mathcal{L}_{\text{sface}}$ is a composite loss that aligns the SMPL-X face geometry (including the personalized offset $\Delta V_{\text{face}}$) to the pre-fitted FLAME model~\cite{Li:FLAME:SA17, Feng:DECA:TOG21}:

\begin{equation}
\mathcal{L}_{\text{sface}} = \mathcal{L}_{\text{vertex}} + \mathcal{L}_{\text{lap}} + \mathcal{L}_{\text{edge}},
\label{eq:sface}
\end{equation}

\noindent
where $\mathcal{L}_{\text{vertex}} = \left\| (\mathbf{V}_{\text{face}} + \Delta \mathbf{V}_{\text{face}}) - \mathbf{V}_{\text{FLAME}} \right\|_1$ is an $L_1$ loss on vertex positions, directly minimizing the distance between the SMPL-X face vertices and the target FLAME vertices to capture fine-grained identity details. $\mathcal{L}_{\text{lap}}$ is an $L_2$ loss on the mesh Laplacians, $\left\| \Delta (\mathbf{V}_{\text{face}} + \Delta \mathbf{V}_{\text{face}}) - \Delta \mathbf{V}_{\text{FLAME}} \right\|_2$, which ensures the facial mesh curvature and local smoothness properties are consistent. $\mathcal{L}_{\text{edge}}$ is an $L_1$ loss on the edge lengths of the two meshes, preserving the local topology and preventing unnatural stretching. Finally, $\mathcal{L}_{\text{sreg}}$ is a regularization term to ensure plausible and stable optimization:

\begin{equation}
\mathcal{L}_{\text{sreg}} = \mathcal{L}_{\text{shape}} + \mathcal{L}_{\text{jo}} + \mathcal{L}_{\text{sym}},
\end{equation}

\noindent
where $\mathcal{L}_{\text{shape}} = \left\| \beta \right\|_2^2$ is a squared $L_2$ norm on the shape parameters $\beta$, keeping the body shape within the plausible distribution of the SMPL-X model. $\mathcal{L}_{\text{jo}} = \left\| \Delta \mathbf{J} \right\|_2^2$ is a squared $L_2$ norm on the joint offsets, preventing extreme and non-human-like deviations in the skeleton. $\mathcal{L}_{\text{sym}}$ enforces bilateral symmetry on the personalized offsets $\Delta \mathbf{J}$ and $\Delta \mathbf{V}_{\text{face}}$ by penalizing differences between corresponding left and right-side offsets. This formulation yields a geometry-consistent and identity-preserving reconstruction that remains robust across varying levels of body visibility. We employ SMPL-X~\cite{Pavlakos:SMPLX:CVPR19} as the unified body representation across the entire visibility spectrum, including head-only inputs, instead of switching to a dedicated facial model such as FLAME~\cite{Li:FLAME:SA17}. This unified formulation enables a single reconstruction pipeline for all input scenarios. To preserve facial fidelity, we align the SMPL-X facial region with a DECA-initialized~\cite{Feng:DECA:TOG21} FLAME mesh using the facial alignment loss $\mathcal{L}_{\text{sface}}$ (Eq.~\ref{eq:sface}) and transfer facial expressions through the shared expression basis.

\vspace{-2mm}
\subsection{Video Generation for Unseen Texture}
\label{subsec:video_generation}
In unconstrained videos with severe cropping or occlusion, existing avatar methods often produce artifacts in regions that are never observed. While some recent works~\cite{lee2024guess,sim2025persona} regularize these unseen areas using diffusion priors, they assume input videos contain full-body coverage, which limits their applicability to partial-body scenarios. To overcome this limitation, we introduce a part-specific motion video generation strategy. Using MimicMotion~\cite{Zhang:MimicMotion:ICML25}, we synthesize auxiliary videos conditioned on the input, and jointly use these with the original footage to construct the final avatar. This augmentation effectively recovers missing texture information for consistently unobserved regions. Specifically, we generate a novel video of the subject performing a full $360^\circ$ rotation. This synthetic motion reveals unobserved views, such as the back, providing the necessary appearance cues for texture completion.
\vspace{-2mm}
\subsection{Representation}
\label{subsec:representation}

We construct our model by integrating the parametric SMPL-X~\cite{Pavlakos:SMPLX:CVPR19} prior with explicit 3D Gaussian splatting~\cite{Kerbl:3DGS:SIGA23}. In line with~\cite{Moon:ExAvatar:ECCV24}, we use a hybrid mesh–Gaussian representation. From the co-registered SMPL-X model (shape $\mathbf{\beta}$, offsets $\Delta \mathbf{J},\Delta \mathbf{V}_{\rm face}$), we generate a canonical mesh $\mathbf{\bar{V}}$ that is up-sampled to $N$ vertices while preserving the underlying triangular connectivity of the SMPL-X template. We associate with each of those $N$ vertices a learnable 3D Gaussian asset (mean position, scale, color), thus forming a surface-conditioned Gaussian representation. To encode identity and enable animation, we use a learnable triplane field $\mathbf{T} \in \mathbb{R}^{3 \times C \times H \times W}$. Canonical positions $\mathbf{\bar{P}} \in \mathbb{R}^{N \times 3}$ are orthogonally projected onto the three planes, and bilinear interpolation yields a per-vertex feature $\mathbf{f}_i$ (\cref{fig:overview_pipeline_figure}). Two MLP groups decode these features. The first predicts pose-independent Gaussian parameters (offset $\Delta \mathbf{V}_{\text{tri},i}$, scale $\mathbf{S}_{\text{tri},i}$, color logit $\mathbf{C}_{\text{tri},i}$) that capture static shape and appearance. The second is pose-dependent: one MLP maps $\mathbf{f}_i$ and body pose (excluding root) to $\Delta \mathbf{V}_{\text{pose},i}, \Delta \mathbf{S}_{\text{pose},i}$, and a color MLP maps pose and normal $\mathbf{n}_i$ to $\Delta \mathbf{C}_{\text{pose},i}$. This disentangles static identity from dynamic articulation, supporting robust learning under limited pose diversity while preserving high-frequency detail.

\vspace{-0.3cm}
\subsection{Visibility-Aware Optimization}
\label{subsec:vis_aware} 
In upper-body or head-only videos, large portions of the body (e.g., legs and feet) are never observed. Methods that assume a full one-to-one correspondence with the SMPL-X surface inevitably hallucinate these unseen regions, leading to unrealistic geometry and texture artifacts in the visible regions. To mitigate this, we introduce a visibility-aware optimization that gathers \emph{per-Gaussian} visibility cues and restricts supervision to only the observed regions. Visibility information, indicating which Gaussians contribute to each rendered pixel, is obtained directly from the rasterizer and used to guide optimization.

\noindent
\textbf{Per-Gaussian Evidence.}
After each refinement stage, we compute a confidence signal for every Gaussian $i$, defined by its visibility rate:

\vspace{-0.3cm}
\begin{align}
  v_i \;=\; \frac{1}{F}\sum_{f=1}^{F} v_i^f \;\in\; [0,\,1],
  \label{eq:vis_rate}
\end{align}
\vspace{-1mm}

\noindent
where $v_i^f$ denotes the binary visibility flag for Gaussian $i$ at frame $f$, and $F$ 
denotes the total number of processed frames.
This score reflects how confidently a Gaussian is supported by visual evidence. We first optimize all Gaussians without filtering for the initial $2{,}000$ iterations. This allows coarse geometry, pose, and color to stabilize, building reliable statistics for $v_i$ and preventing early false negatives from aggressively removing still-emerging regions. After warm-up, we determine the visibility threshold automatically via Otsu's method~\cite{otsu1979threshold}: the distribution of $\{v_i\}$ across all Gaussians is treated as a bimodal histogram, and the threshold $\tau^*$ that minimizes intra-class variance between the visible and invisible Gaussians is selected. This avoids the need for a manually tuned threshold and adapts to the observed visibility distribution of each input sequence. Gaussians whose visibility score falls below the threshold $\tau^*$ are excluded from the forward rasterization pass, reducing computational overhead in proportion to the number of invisible Gaussians. This visibility-based filtering is similarly applied to the regularization terms, specifically the Laplacian constraints on Gaussian color, scale, and position: these are masked for invisible Gaussians, preventing them from contributing to the loss and ensuring that regularization does not introduce artifacts into the visible regions. 

\begin{wrapfigure}[9]{r}{0.50\textwidth}
  \centering
  \vspace{-0.35cm}
  \includegraphics[width=0.5\textwidth]{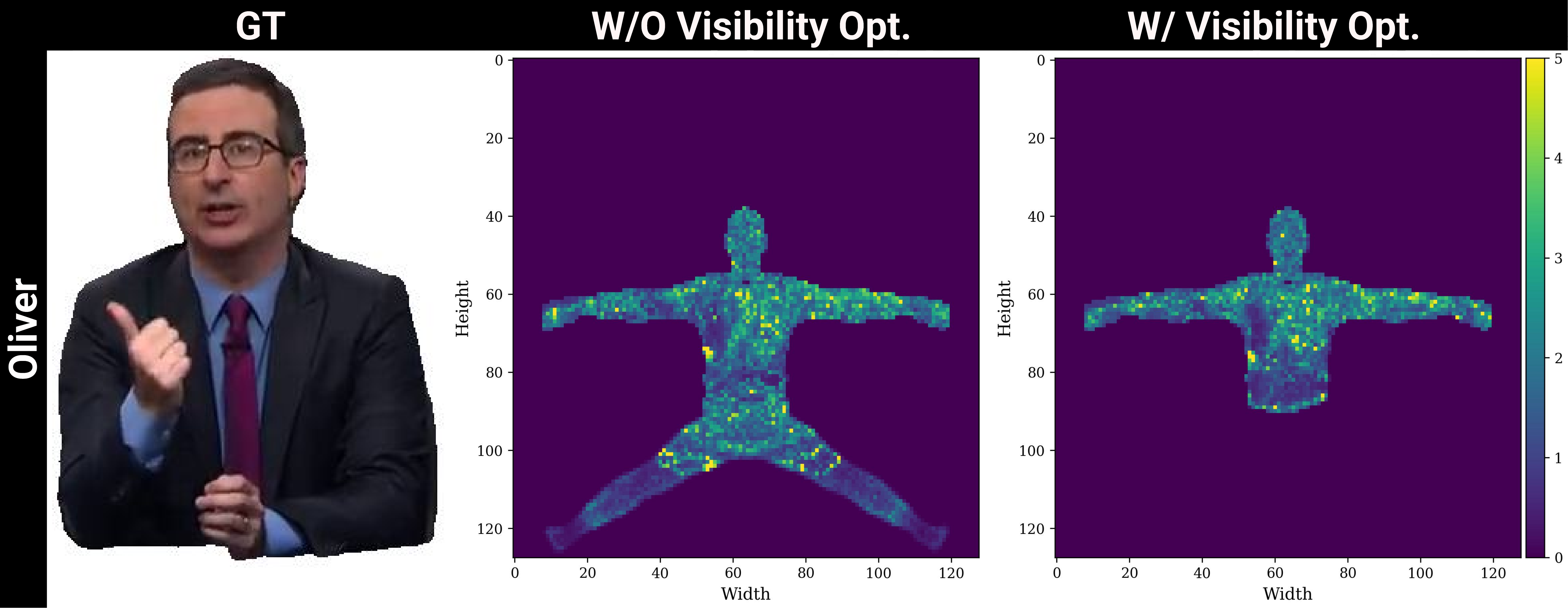}
  \caption{Gradient magnitude on the shared triplane for the
    upper-body \textit{Oliver} sequence.}
  \label{fig:grad}
\end{wrapfigure}

Unlike conventional visibility masks, $v_i$ is used to guide optimization rather than as a rendering-time filter. Since all Gaussians are decoded from the shared triplane representation $T$, gradients originating from observed pixels can propagate to Gaussians in unobserved regions and, through the shared representation, indirectly influence neighboring visible Gaussians. By masking invisible Gaussians in both the reconstruction and regularization losses (Sec.~3.4), we suppress these spurious gradients at their source, ensuring that the optimization is driven only by regions supported by visual evidence. As illustrated in Fig.~\ref{fig:grad}, on the \textit{Oliver} sequence, where the lower limbs are never observed, conventional optimization propagates gradients into the invisible lower body, which subsequently degrades the reconstructed torso. In contrast, our visibility-aware optimization confines gradient propagation to the observed upper body, resulting in more stable and accurate reconstruction.

\vspace{-1.0em}
\subsection{Part-Specific Residual Refinement}
\label{subsec:refinement}
Capturing fine-grained facial expressions and accurate hand details is crucial for conveying emotion and maintaining realism~\cite{Jeong_2026_CVPR,sayem2026handvqa}. To enhance these high-frequency regions, we introduce an implicit function–based residual appearance refinement module that enhances color predictions for the face and hands~\cite{haider2026inr}. 
Given a Gaussian center $\boldsymbol{\mu} \in \mathbb{R}^3$ , time $t$, and pose $\theta$, we construct the input by concatenating their positional encodings. A lightweight MLP then predicts a color residual conditioned on this spatiotemporal encoding, allowing precise recovery of dynamic appearance details beyond the base Gaussian representation.
\vspace{-0.2cm}
\begin{equation}
\Delta \mathbf{C}_{\text{res}}(\boldsymbol{\mu}, t,\theta\big)
\;=\;
F_{\theta}\!\big(\mathrm{concat}\big(\gamma(\boldsymbol{\mu}),\, \gamma(t)\big) , \theta\big )
\end{equation}
where $\Delta \mathbf{C}_{\text{res}}$ represents the predicted color residuals, $F_{\theta}$ denotes the learnable MLP, and $\gamma(\cdot)$ is the positional encoding function. In practice, we implement $F_{\theta}$ using three separate, lightweight MLPs 
specialized for the face, left hand, and right hand, respectively, each of which 
is only activated when the corresponding body part is visible according to our 
visibility-aware optimization (Sec.~\ref{subsec:vis_aware}). We first train the coarse avatar without these residual modules for $2{,}000$ iterations to obtain a stable geometry and base color representation. Once convergence is reached, the three residual networks are activated, and their predicted color offsets are progressively added to enhance fine-grained appearance. The final color is then updated as follows:
\begin{equation}
\mathbf{C}_{\text{final}} \;=\; \mathbf{C}_{\text{tri}} \;+\; \Delta \mathbf{C}_{\text{pose}} \;+\; \Delta \mathbf{C}_{\text{res}}.
\end{equation}

\subsection{Animation and Rendering}
\label{subsec:animcation_rendering}

The 3D human avatar is constructed in a canonical space and animated using SMPL-X poses $\theta$ and facial expression code $\psi$. Identity-specific offsets $\Delta \mathbf{V}_{\text{tri}}$ are combined with pose-dependent offsets $\Delta \mathbf{V}_{\text{pose}}$, while face and hand regions directly adopt SMPL-X deformations for accuracy. Facial expression offsets $\Delta \mathbf{V}_{\text{expr}}$ from $\psi$ are added to face vertices, eliminating the need to learn the expression space. The posed geometry is then skinned via linear blend skinning (LBS) using SMPL-X weights. Rendering is performed using 3D Gaussian Splatting (3DGS)~\cite{Kerbl:3DGS:SIGA23}.

\noindent
\textbf{Training Objective.}
We jointly optimize all learnable components, including the 3D Gaussian parameters
(positions, scales, colors, and opacities), triplane feature fields, residual RGB MLPs, alongside the sequence-specific SMPL-X parameters.
We supervise the framework using a composite image reconstruction objective comprising
$L_1$, SSIM, and perceptual LPIPS terms between rendered outputs and ground-truth images.
We further incorporate a dedicated facial
consistency loss and apply Laplacian-style regularization following ~\cite{Moon:ExAvatar:ECCV24} to encourage spatial smoothness
in geometry and appearance.
During mixed-source training, captured and diffusion-generated frames are sampled with
equal probability; to account for potential artifacts in synthetic views, we down-weight
their reconstruction losses following~\cite{sim2025persona}.
Formally, the overall training objective $\mathcal{L}$ is a weighted sum defined as:

\begin{equation}
\begin{aligned} 
\mathcal{L}
= &\;
\lambda_{L1}\,\mathcal{L}_{1}
+ \lambda_{\text{ssim}}\,\mathcal{L}_{\text{ssim}}
+ \lambda_{\text{lpips}}\,\mathcal{L}_{\text{lpips}}
+ \lambda_{\text{face}}\,\mathcal{L}_{\text{face}}
+ \lambda_{\text{reg}}\,\mathcal{L}_{\text{reg}}
\end{aligned}
\label{eq:main_loss}
\end{equation}

\noindent
where $\mathcal{L}_{1}$, $\mathcal{L}_{\text{ssim}}$, and $\mathcal{L}_{\text{lpips}}$ are computed on a cropped human region for efficiency. The $L_{1}$ term penalizes per-pixel deviations between the rendered image $I_{\text{pred}}$ and the ground-truth $I_{\text{gt}}$, $\mathcal{L}_{\text{ssim}}$ encourages local structural and contrast consistency, and $\mathcal{L}_{\text{lpips}}$ compares deep feature activations to promote sharp, perceptually realistic textures. The facial loss $\mathcal{L}_{\text{face}}$ enforces that facial Gaussians remain aligned with a stable, identity-preserving facial template. Following~\cite{Moon:ExAvatar:ECCV24}, we render a facial image $I{\text{ mesh}}$ using a differentiable mesh renderer and a fixed UV texture obtained by averaging registration textures, and supervise it with
$\mathcal{L}_{\text{face}} = \bigl\| I_{\text{mesh}} - I_{\text{gt}} \bigr\|_1$
over the facial region. This constrains the facial Gaussians to maintain a consistent appearance and reduces drift under novel expressions. 

The regularization term $\mathcal{L}_{\text{reg}}$ aggregates our geometry and appearance priors. Its main components are Laplacian smoothness penalties, which we apply in a visibility-aware manner. Using the binary visibility mask $\mathbf{m}_i \in \{0,1\}$ (derived from the visibility rate $v_i$ in~\cref{subsec:vis_aware}), we ensure that smoothing is applied only to well-observed Gaussians, preventing over-regularization of unseen regions.

\section{Experiments}
\label{sec:experiments}
We present empirical results on standard benchmarks. We first cover implementation details, datasets, and metrics (\cref{sec:implementation}), then quantitative and qualitative comparisons (\cref{sec:quant},~\cref{sec:qual}), and finally an ablation analysis (\cref{sec:ablations}).

\vspace{-1em}
\subsection{Implementation Details}
\label{sec:implementation}
Our framework is implemented in PyTorch~\cite{Paszke:PyTorch:NIPS19} and optimized using the Adam optimizer~\cite{Kingma:Adam:ICLR15}. All experiments are conducted on a single NVIDIA RTX A6000 GPU for 30k iterations with a batch size of 1. The base learning rate for all network parameters is set to $1\times10^{-3}$. Body masks are extracted using SAM~\cite{SegmentAnything}. To reduce sensitivity to imperfect foreground masks, we use the same masking strategy as~\cite{Moon:ExAvatar:ECCV24}. Detailed hyperparameter settings, including the $\lambda$ weights used in~\cref{eq:tracking_loss} and~\cref{eq:main_loss}, are provided in the Supplementary Material.

\noindent
\textbf{Datasets.}
\label{sec:dataset}
We evaluate our method across varying visibility inputs, (i) full-body, (ii) upper-body, and (iii) head-only, to comprehensively evaluate the performance of FlexiAvatar. For full-body evaluation, we use the NeuMan dataset~\cite{Jiang:NeuMan:ECCV22}, selecting the \texttt{bike}, \texttt{citron}, \texttt{jogging}, and \texttt{seattle} sequences, which provide broad body coverage with minimal motion blur. We adopt the official train/test split following the established protocol~\cite{Hu:GaussianAvatar:CVPR24}. To further validate generalization on a standard full-body benchmark, we additionally evaluate on ZJU-MoCap~\cite{zjumocap_dataset} using six subjects (377, 386, 387, 392, 393, 394). For each subject, we reserve the last 100 frames for testing and utilize the remaining frames for training. 
To further assess robustness in unconstrained real-world scenarios, we evaluate our method on WildAvatar~\cite{wildavatar:CVPR2025}, a large-scale human-centric dataset collected from YouTube. We randomly select seven subjects, reserving the last 100 frames of each sequence for testing and using the remaining frames for training.
For upper-body evaluation, we use three subjects (\texttt{Oliver}, \texttt{Conan}, and \texttt{Chemistry}) from the TalkShow dataset~\cite{Yi:TalkShow:CVPR23}. Each 10-second clip ($\sim$300 frames) is split by reserving the last 30 frames for testing and using the remaining frames for training. Finally, for head-only reconstruction, we evaluate on seven videos (\texttt{bala}, \texttt{malte\_1}, \texttt{biden}, \texttt{justin}, \texttt{marcel}, \texttt{nf\_01}, \texttt{nf\_03}) from the INSTA dataset~\cite{INSTA:CVPR23}. Following prior work, we reserve the final 350 frames for testing~\cite{RGBAvatar:CVPR25}.

\begin{table*}[t]
\parbox{.485\linewidth}{
    \centering
    \captionsetup{width=\linewidth}
    \caption{Quantitative results on the NeuMan~\cite{Jiang:NeuMan:ECCV22} dataset for 3D human avatar reconstruction.}
    \label{tab:avg_combined_neuman}
    \vspace{-3mm}
    \resizebox{0.85\linewidth}{!}{%
    \begin{tabular}{@{}l|ccc@{}}
    \toprule
    Method & PSNR$\uparrow$ & SSIM$\uparrow$ & LPIPS$\downarrow$ \\
    \midrule
    HumanNeRF~\cite{Weng:HumanNeRF:CVPR22} & 27.06 & 0.967 & 1.90 \\
    InstantAvatar~\cite{Jiang:InstantAvatar:CVPR23} & 28.47 & 0.972 & 2.80 \\
    NeuMan~\cite{Jiang:NeuMan:ECCV22} & 29.32 & 0.972 & 1.40 \\
    Vid2Avatar~\cite{Guo:Vid2Avatar:CVPR23} & 30.70 & 0.980 & 1.40 \\
    GaussianAvatar~\cite{Hu:GaussianAvatar:CVPR24} & 29.94 & 0.980 & 1.20 \\
    3DGS-Avatar~\cite{Qian:3DGSAvatar:CVPR24} & 28.99 & 0.974 & 1.60 \\
    ExAvatar~\cite{Moon:ExAvatar:ECCV24} & 34.80 & 0.984 & 0.90 \\
    Vid2AvatarPro~\cite{Guo:Vid2AvatarPro:CVPR25} & 32.71 & 0.983 & 1.19 \\
    \midrule
    \textbf{Ours} & \textbf{35.77} & \textbf{0.987} & \textbf{0.83} \\
    \bottomrule
    \end{tabular}%
    }
}
\hfill
\parbox{.48\linewidth}{
    \centering
    \captionsetup{width=\linewidth}
    \caption{ZJU-MoCap dataset results (subjects 377, 386, 387, 392, 393, 394).}
    \label{tab:zju_mocap}
    \vspace{-3mm}
    \resizebox{0.85\linewidth}{!}{%
    \begin{tabular}{@{}lcccccc@{}}
    \toprule
    Method & 377 & 386 & 387 & 392 & 393 & 394 \\
    \midrule
    \multicolumn{7}{@{}c}{\textbf{PSNR}$\uparrow$} \\
    \midrule
    GauHuman~\cite{gauhuman:CVPR2024} & 32.63 & 36.47 & 32.99 & 32.47 & 32.31 & 33.81 \\
    ToMiE~\cite{tomie:ICCV2025}       & 33.63 & 37.29 & 34.42 & 34.46 & 32.73 & 34.26 \\
    \textbf{Ours}                    & \textbf{34.98} & \textbf{39.50} & \textbf{36.11} & \textbf{36.15} & \textbf{34.66} & \textbf{36.63} \\
    \midrule
    \multicolumn{7}{@{}c}{\textbf{SSIM}$\uparrow$} \\
    \midrule
    GauHuman~\cite{gauhuman:CVPR2024} & 0.974 & 0.973 & 0.970 & 0.967 & 0.965 & 0.967 \\
    ToMiE~\cite{tomie:ICCV2025}       & 0.977 & 0.976 & 0.973 & 0.973 & 0.966 & 0.968 \\
    \textbf{Ours}                    & \textbf{0.991} & \textbf{0.990} & \textbf{0.986} & \textbf{0.991} & \textbf{0.985} & \textbf{0.986} \\
    \midrule
    \multicolumn{7}{@{}c}{\textbf{LPIPS}$\downarrow$} \\ 
    \midrule
    GauHuman~\cite{gauhuman:CVPR2024} & 2.07 & 2.99 & 2.75 & 3.35 & 3.14 & 2.57 \\
    ToMiE~\cite{tomie:ICCV2025}       & 1.80 & 2.91 & 2.50 & 2.88 & 3.02 & 2.67 \\
    \textbf{Ours}                    & \textbf{0.66} & \textbf{0.85} & \textbf{0.99} & \textbf{0.80} & \textbf{1.15} & \textbf{1.01} \\
    \bottomrule
    \end{tabular}%
    }
}
\end{table*}

\begin{table*}[t]
\parbox{.455\linewidth}{
    \centering
    \captionsetup{width=\linewidth}
    \caption{Quantitative comparisons of 3D human avatars on the TalkShow (upper-body)~\cite{Yi:TalkShow:CVPR23} dataset.}
    \label{tab:talkshow_multi}
    \vspace{-3mm}
    \resizebox{0.85\linewidth}{!}{%
    \begin{tabular}{@{}lc|ccc@{}}
    \toprule
    Subject & Method & PSNR$\uparrow$ & SSIM$\uparrow$ & LPIPS$\downarrow$ \\
    \midrule
    \multirow{4}{*}{Oliver}
    & ExAvatar~\cite{Moon:ExAvatar:ECCV24} & 29.13 & 0.938 & 2.29 \\
    & GART~\cite{Lei:GART:CVPR24}         & 24.14 & 0.927 & 7.76 \\
    & GUAVA~\cite{Zhang:GUAVA:ICCV25}     & 26.76 & 0.890 & 12.17 \\
    & \textbf{Ours}                       & \textbf{29.80} & \textbf{0.952} & \textbf{1.81} \\
    \midrule
    \multirow{4}{*}{Conan}
    & ExAvatar~\cite{Moon:ExAvatar:ECCV24} & 35.66 & 0.980 & 3.61 \\
    & GART~\cite{Lei:GART:CVPR24}         & 26.89 & 0.974 & 5.56 \\
    & GUAVA~\cite{Zhang:GUAVA:ICCV25}     & 29.70 & 0.928 & 7.69 \\
    & \textbf{Ours}                       & \textbf{36.73} & \textbf{0.984} & \textbf{2.82} \\
    \midrule
    \multirow{4}{*}{Chemistry}
    & ExAvatar~\cite{Moon:ExAvatar:ECCV24} & 25.67 & 0.922 & 10.29 \\
    & GART~\cite{Lei:GART:CVPR24}         & 22.28 & 0.914 & 11.89 \\
    & GUAVA~\cite{Zhang:GUAVA:ICCV25}     & 26.85 & 0.920 & \textbf{7.15} \\
    & \textbf{Ours}                       & \textbf{27.80} & \textbf{0.935} & 9.67 \\
    \bottomrule
    \end{tabular}%
    }
}
\hfill
\parbox{.51\linewidth}{
    \centering
    \captionsetup{width=\linewidth}
    \caption{Quantitative comparisons of 3D human avatars on the INSTA (Head-only)~\cite{INSTA:CVPR23} dataset.}
    \label{tab:insta_avg_combined}
    \vspace{-2.5mm}
    \resizebox{0.9\linewidth}{!}{%
    \begin{tabular}{@{}l|ccc@{}}
    \toprule
    Method & PSNR$\uparrow$ & SSIM$\uparrow$ & LPIPS$\downarrow$ \\
    \midrule
    SplattingAvatar~\cite{Shao:SplattingAvatar:CVPR24} & 29.03 & 0.932 & 10.35 \\
    GaussianAvatars~\cite{Qian:GaussianAvatars:CVPR24} & 29.10 & 0.945 & 8.61 \\
    FlashAvatar~\cite{Xiang:FlashAvatar:CVPR24} & 27.44 & 0.912 & 11.05 \\
    MonoGaussianAvatar~\cite{Chen:MonoGaussianAvatar:SIGGRAPH24} & 28.91 & 0.945 & 7.43 \\
    GaussianBlendShapes~\cite{Ma:GaussianBlendshapes:TOG24} & 30.01 & 0.947 & 9.17 \\
    FATE~\cite{Zhang:FATE:CVPR25} & 27.85 & 0.942 & 5.68 \\
    RGBAvatar~\cite{RGBAvatar:CVPR25} & 32.72 & \textbf{0.953} & 6.04 \\
    \midrule
    \textbf{Ours} & \textbf{33.04} & \textbf{0.953} & \textbf{5.63} \\
    \bottomrule
    \end{tabular}%
    }
}
\vspace{-2mm}
\end{table*}

\begin{figure}[t]
\centering
\footnotesize
\renewcommand{\arraystretch}{1.1}
\begin{minipage}[t]{0.46\linewidth}
\centering
\vspace{-70pt}
\captionof{table}{Quantitative comparisons of 3D human avatars on the WildAvatar~\cite{wildavatar:CVPR2025} dataset.}
\label{tab:wildavatar}
\setlength{\tabcolsep}{4pt}
\scriptsize
\begin{tabular}{@{}lccc@{}}
\toprule
Method & PSNR$\uparrow$ & SSIM$\uparrow$ & LPIPS$\downarrow$ \\
\midrule
GauHuman~\cite{gauhuman:CVPR2024}  & 28.31 & 0.957 & 3.83 \\
ToMiE~\cite{tomie:ICCV2025}    & 28.65 & 0.957 & 3.81 \\
\textbf{Ours} & \textbf{30.98} & \textbf{0.973} & \textbf{2.93} \\
\bottomrule
\end{tabular}
\end{minipage}
\hfill
\begin{minipage}[t]{0.50\linewidth}
\centering
\captionsetup{font=scriptsize} 
\includegraphics[width=\linewidth]{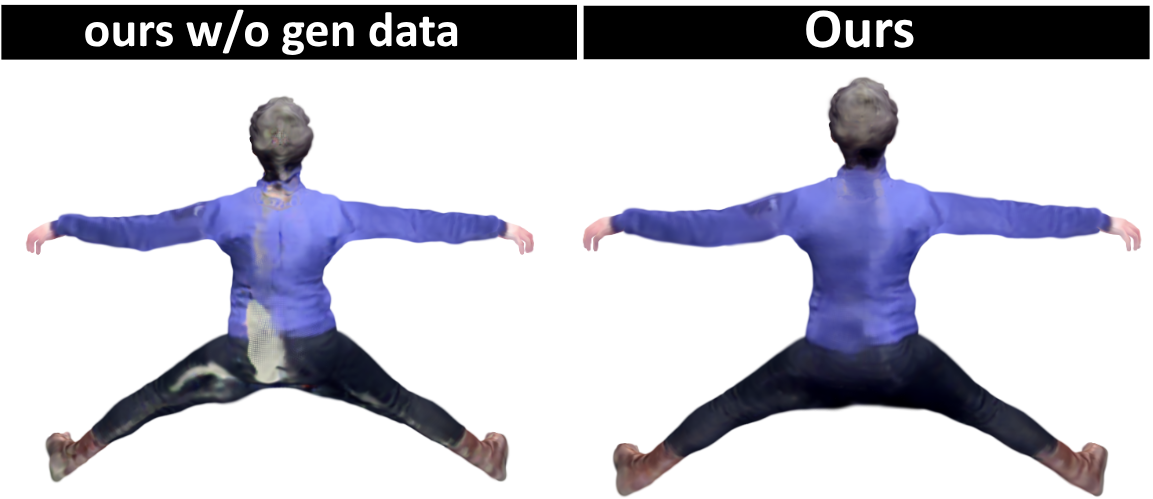}
\caption{Avatar in canonical space with and without generated data in our method. Generated data fills missing texture without noticeable artifacts.}
\label{fig:canonical_abalation}
\end{minipage}
\end{figure}

\noindent
\textbf{Evaluation Protocol.}
\label{sec:protocol}
For quantitative evaluation, we report PSNR, SSIM~\cite{Wang:SSIM}, and LPIPS ($\times$100)~\cite{zhang:LPIPS}. In all result tables, the best score is highlighted in \textbf{bold}. For~\cref{tab:avg_combined_neuman}, all baseline numbers are taken directly from the corresponding papers~\cite{Hu:GaussianAvatar:CVPR24, Guo:Vid2AvatarPro:CVPR25, Moon:ExAvatar:ECCV24}. Consistent with prior work~\cite{animatableneuralradiance, Hu:GaussianAvatar:CVPR24, Qian:3DGSAvatar:CVPR24, Jiang:InstantAvatar:CVPR23}, evaluation on NeuMan is performed by optimizing SMPL-X parameters for test frames using only the image loss while keeping all other model parameters fixed. 
For quantitative evaluation, we follow the background masking strategy of ExAvatar~\cite{Moon:ExAvatar:ECCV24}.
For~\cref{tab:talkshow_multi}, we report metrics computed using the official implementations released by the respective methods. For the ZJU-MoCap benchmark, quantitative results for GauHuman~\cite{gauhuman:CVPR2024} and ToMiE~\cite{tomie:ICCV2025} were obtained by executing their officially released source code on our exact 100-frame test split to ensure a fair comparison. For~\cref{tab:insta_avg_combined}, baseline numbers are adopted from ~\cite{RGBAvatar:CVPR25, Zhang:FATE:CVPR25} , with RGBAvatar~\cite{RGBAvatar:CVPR25} computed from its official code.

\vspace{-1em}
\subsection{Quantitative Comparison}
\begin{figure}[t!]
\centering
\includegraphics[width=\linewidth]{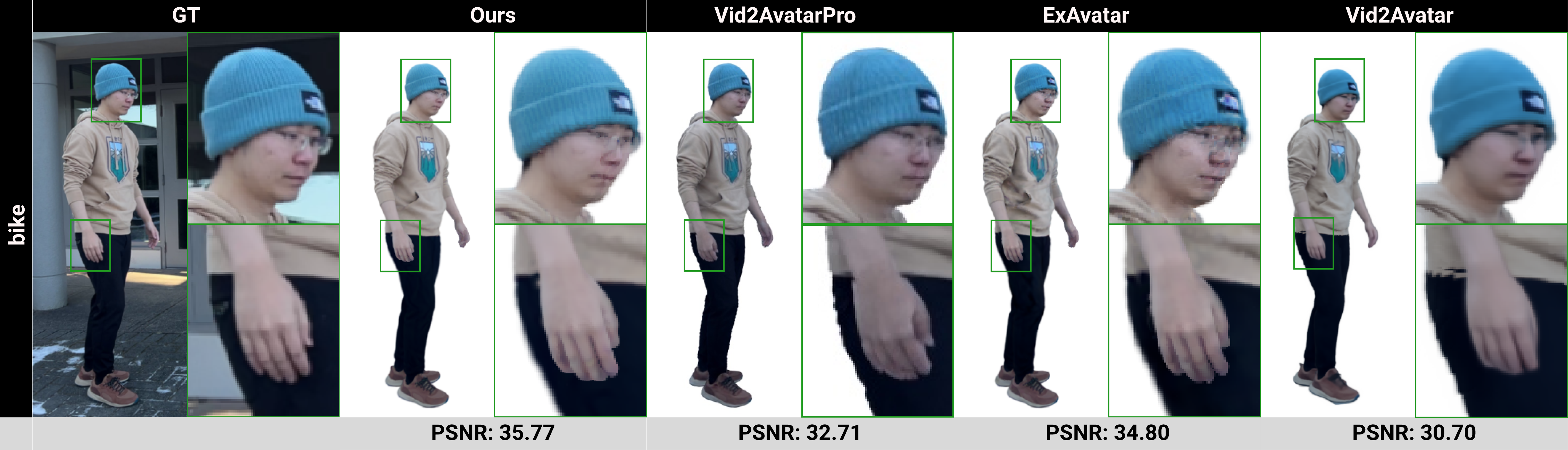}
\caption{Qualitative comparison on the NeuMan full-body dataset. Our method recovers 
sharper clothing texture and more accurate hand detail.  }
\label{fig:qual_neuman}
\end{figure}
\begin{figure}[t!]
\centering
\includegraphics[width=\linewidth]{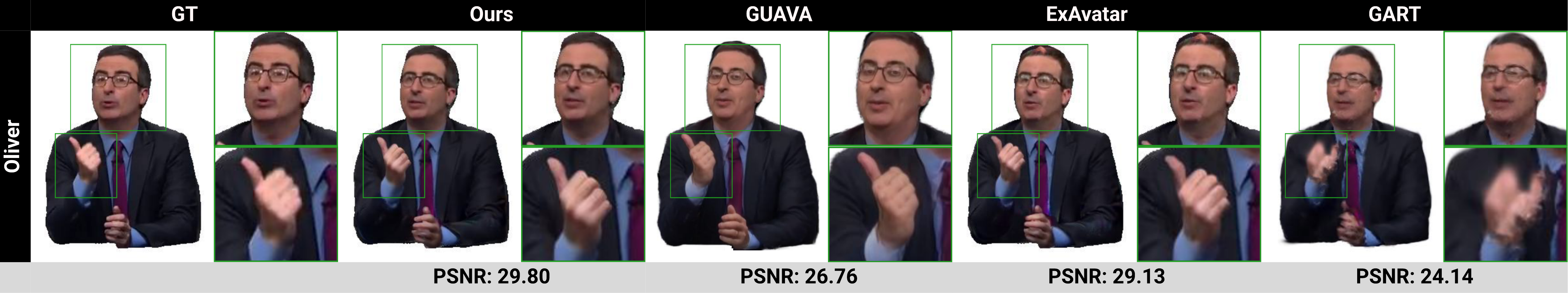}
\caption{Qualitative comparison on the TalkShow upper-body dataset. Our method preserves facial sharpness and recovers hand detail.}
\label{fig:qual_talkshow}
\end{figure}
\label{sec:quant}
We conduct a comprehensive evaluation against SOTA methods across three settings: full-body (NeuMan, ZJU-MoCap), upper-body (TalkShow), and head-only (INSTA).
  
\noindent
\textbf{Full-Body Avatars on NeuMan.}
We evaluate our method against state-of-the-art full-body avatar approaches on the NeuMan dataset~\cite{Jiang:NeuMan:ECCV22}. As reported in~\cref{tab:avg_combined_neuman}, our model achieves the highest performance across all evaluation metrics. To further assess the fidelity of fine-grained reconstruction, we conduct focused evaluations on the facial and hand regions that exhibit complex, high-frequency motion and appearance variations. As shown in the Supplementary Material (Sec. E.6, Tab. A7), the proposed part-specific residual refinement (Sec.~3.5) significantly improves the reconstruction of fine-grained facial and hand details, consistently outperforming ExAvatar on cropped face and hand regions with notably higher PSNR. These results demonstrate the effectiveness of our framework in recovering detailed and expressive geometry and appearance.

\noindent
\textbf{Full-Body Avatars on ZJU-MoCap.}
We further validate our method on the ZJU-MoCap dataset~\cite{zjumocap_dataset}, 
a standard full-body benchmark, comparing against GauHuman~\cite{gauhuman:CVPR2024} 
and ToMiE~\cite{tomie:ICCV2025} across six subjects (377, 386, 387, 392, 393, 394). 
As reported in~\cref{tab:zju_mocap}, our method achieves the best performance across all subjects and all three metrics, outperforming both baselines especially in LPIPS. This confirms that our visibility-aware optimization preserves full-body reconstruction quality while generalizing across constrained studio captures and partial-visibility settings within one pipeline.

\noindent
\textbf{Full-Body Avatars on WildAvatar.} To evaluate generalization beyond curated benchmarks, we compare FlexiAvatar against GauHuman~\cite{gauhuman:CVPR2024} and ToMiE~\cite{tomie:ICCV2025} on seven in-the-wild subjects from the WildAvatar dataset~\cite{wildavatar:CVPR2025}. As reported in \cref{tab:wildavatar}, FlexiAvatar consistently outperforms both baselines across all three metrics, achieving higher PSNR and lower LPIPS. These results demonstrate that our visibility-aware optimization generalizes effectively to unconstrained, in-the-wild videos.

\noindent
\textbf{Upper-Body Avatars on TalkShow.}
We also evaluate our method in the upper-body setting using three subjects from the TalkShow dataset. As shown in~\cref{tab:talkshow_multi}, our approach consistently outperforms all baselines (ExAvatar~\cite{Moon:ExAvatar:ECCV24}, GART~\cite{Lei:GART:CVPR24}, GUAVA~\cite{Zhang:GUAVA:ICCV25}) across all subjects. This confirms the robustness and high fidelity of our method for upper-body reconstruction, particularly in challenging regions with limited visibility. The effectiveness of our framework in recovering detailed and expressive geometry and appearance further highlights its ability to generalize across varying input coverage, maintaining high-frequency details even when only partial observations are available.

\noindent
\textbf{Head-Only Avatars on INSTA.}
Finally, we evaluate our model on head-only reconstruction using seven subjects from the INSTA dataset~\cite{INSTA:CVPR23}. The results are reported in~\cref{tab:insta_avg_combined}.
As evident, our method achieves the best PSNR and LPIPS scores, highlighting its strong generalization ability for dynamic head reconstruction. These results further demonstrate that our visibility-aware design and high-frequency residual modeling remain effective even under extremely limited input coverage.

\vspace{-1em}
\subsection{Qualitative Comparison}
\label{sec:qual}
For qualitative evaluation, we compare FlexiAvatar with a wide range of state-of-the-art methods under three visibility settings: full-body, upper-body, and head-only inputs. For full-body reconstruction, we compare against Vid2Avatar, Vid2AvatarPro, and ExAvatar on NeuMan (Fig.~\ref{fig:qual_neuman}), and against GauHuman and ToMiE on ZJU-MoCap. Our method consistently reconstructs sharper clothing textures and more accurate hand structures on NeuMan, while producing clearer skin textures and more consistent limb geometry on ZJU-MoCap (see Fig.~A3 in the Supplementary Material).
For upper-body reconstruction (Fig.~\ref{fig:qual_talkshow}), we compare with GUAVA, ExAvatar, and GART on the TalkShow dataset. Methods designed under the assumption of full-body visibility degrade under partial observations: GART produces severe artifacts and distorted textures, while ExAvatar yields blurred appearance and fails to capture high-frequency facial and hand details. In contrast, FlexiAvatar preserves facial sharpness and reconstructs fine hand structures, outperforming both general and upper-body-specific baselines such as GUAVA. For head-only reconstruction (Fig.~\ref{fig:main_teaser}), we additionally compare against RGBAvatar, where FlexiAvatar maintains high visual fidelity and produces clean avatar reconstructions, demonstrating strong generalization even under extremely limited input coverage.

\vspace{-2mm}
\subsection{Ablation Studies}
\label{sec:ablations}
To assess the contribution of each core component, we conduct ablation studies on both the NeuMan and INSTA datasets. The quantitative results are reported in~\cref{tab:neuman_ablation} and~\cref{tab:insta_ablation_avg}.
Additional experiments and visualizations are provided in the Supplementary Material, including implementation details, network architectures, and the full SMPL-X registration procedure. We also present extended qualitative comparisons, canonical-space visualizations, and additional novel-pose animation results.

\noindent
\textbf{w/o visibility-aware optimization.}
Disabling our visibility-aware optimization on the INSTA dataset leads to a consistent performance drop across all metrics (\cref{tab:insta_ablation_avg}). The reductions in PSNR and LPIPS confirm that this component is essential for achieving accurate, high-fidelity head reconstruction in monocular, head-only scenarios where visible evidence is limited. Beyond preserving visual fidelity, this visibility-aware pruning inherently restricts the total number of Gaussians that must be optimized and rendered. As detailed in Tab.~\ref{tab:runtime_combined}, this structural efficiency yields a nearly 50\% reduction in memory footprint for head-only avatars, resulting in notably faster animation and rendering speeds compared to full-body baselines.

\begin{table*}[t]
\centering
\parbox{.46\linewidth}{
\centering
\captionsetup{width=\linewidth}
\caption{Impact of refinement heads and generated views on reconstruction quality on NeuMan~\cite{Jiang:NeuMan:ECCV22} dataset.}
\label{tab:neuman_ablation}
\vspace{-3mm}
\resizebox{0.9\linewidth}{!}{
\begin{tabular}{@{}lccc@{}}
\toprule
Method & PSNR$\uparrow$ & SSIM$\uparrow$ & LPIPS$\downarrow$ \\
\midrule
\textbf{Ours (full)} & \textbf{35.77} & \textbf{0.987} & \textbf{0.83} \\
w/o refinement & 34.90 & 0.985 & 0.86 \\
w/o gen. data & 35.29 & 0.986 & 0.92 \\
\bottomrule
\end{tabular}
}
}
\hfill
\parbox{.52\linewidth}{
\centering
\captionsetup{width=\linewidth}
\caption{Impact of visibility-aware optimization and SMPL-X optimization on reconstruction quality for the INSTA~\cite{INSTA:CVPR23} dataset.}
\label{tab:insta_ablation_avg}
\vspace{-3mm}
\resizebox{0.9\linewidth}{!}{
\begin{tabular}{@{}lccc@{}}
\toprule
Method & PSNR$\uparrow$ & SSIM$\uparrow$ & LPIPS$\downarrow$ \\
\midrule
\textbf{Ours (full)} & \textbf{33.04} & \textbf{0.953} & \textbf{5.63} \\
w/o visibility-aware opt. & 32.15 & 0.949 & 5.80 \\
w/o SMPL-X opt. & 31.07 & 0.934 & 7.60 \\
\bottomrule
\end{tabular}
}
}

\vspace{-2mm}
\end{table*}
\begin{table}[t]
\centering
\normalsize 
\setlength{\tabcolsep}{3pt}
\renewcommand{\arraystretch}{1.00}
\caption{
Detailed runtime and memory statistics averaged for Head-only (INSTA) and Upper-body (TalkShow) settings.
\#G denotes the number of human Gaussians.
Asset (MB) is the summed tensor footprint of the Gaussian asset dictionary
(mean, opacity, scale, rotation, rgb). Animation (Anim) and rendering (Rend) speeds are reported in milliseconds.}
\label{tab:runtime_combined}
\vspace{-3mm}
\resizebox{\linewidth}{!}{
\begin{tabular}{llccccccc}
\toprule
Setting & Method & Resolution & \#G & Asset (MB) & Anim (ms) & Rend (ms) & Total (ms) & FPS $\uparrow$ \\
\midrule
\multirow{2}{*}{Head}
& ExAvatar
& $512^2$
& 167{,}390 & 8.94 & 32.175 & 2.098 & 34.272 & 29.18 \\

& Ours
& $512^2$
& \textbf{84{,}095} {\scriptsize(\highlight{-49.7\%})}
& \textbf{4.49} {\scriptsize(\highlight{-49.8\%})}
& \textbf{24.135} {\scriptsize(\highlight{-25.0\%})}
& \textbf{1.410} {\scriptsize(\highlight{-32.8\%})}
& \textbf{25.545} {\scriptsize(\highlight{-25.5\%})}
& \textbf{39.15} {\scriptsize(\highlight{+34.2\%})} \\

\midrule

\multirow{2}{*}{Upper body}
& ExAvatar
& $512^2$
& 167{,}390 & 8.94 & 30.239 & 1.966 & 32.206 & 31.06 \\

& Ours
& $512^2$
& \textbf{144{,}350} {\scriptsize(\highlight{-13.8\%})}
& \textbf{7.71} {\scriptsize(\highlight{-13.8\%})}
& \textbf{28.127} {\scriptsize(\highlight{-7.0\%})}
& \textbf{1.408} {\scriptsize(\highlight{-28.4\%})}
& \textbf{29.536} {\scriptsize(\highlight{-8.2\%})}
& \textbf{33.87} {\scriptsize(\highlight{+9.1\%})} \\

\hline
\end{tabular}
}
\end{table}
\begin{figure}[t]
\centering

\begin{minipage}[t]{0.47\linewidth}
    \centering
    \includegraphics[width=0.95\linewidth]{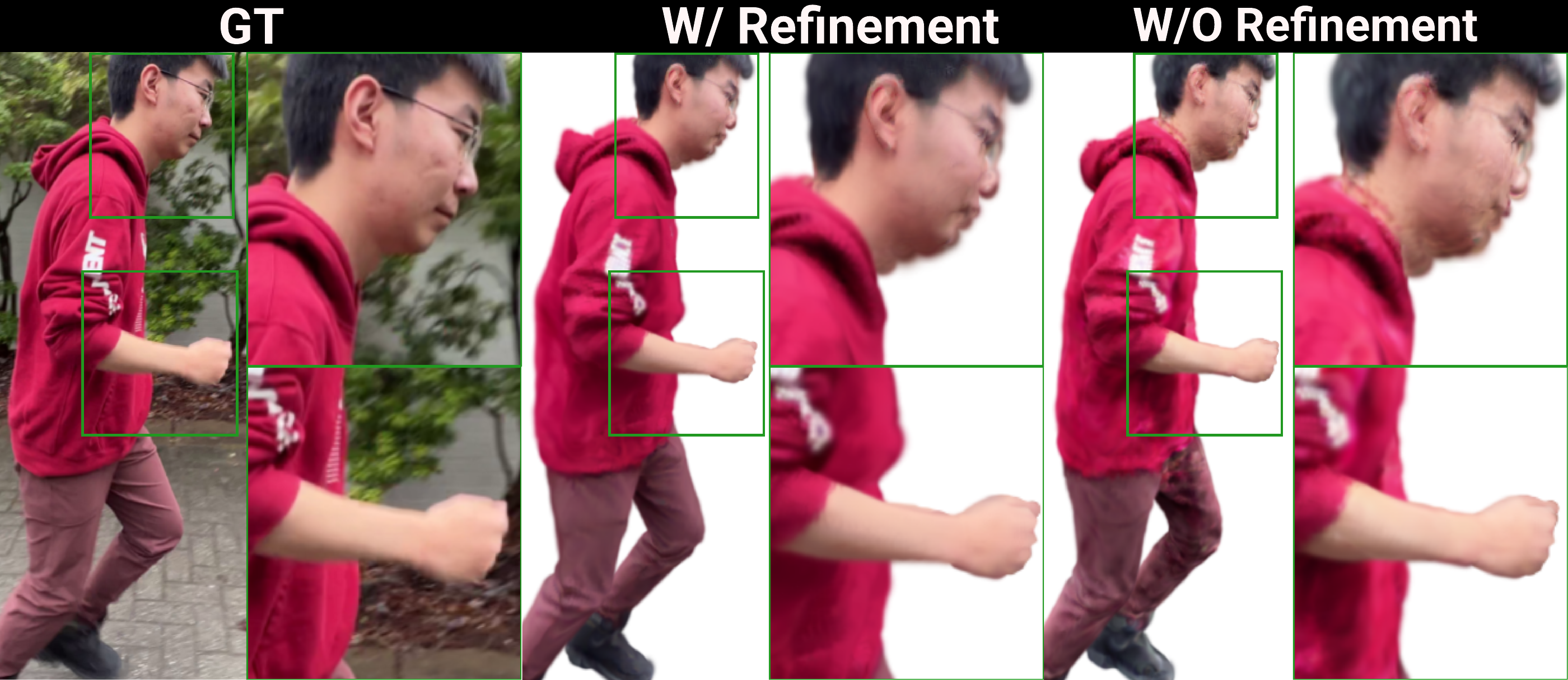}
    \caption{Ablation of the part-specific residual refinement module on 
the NeuMan sequence. FlexiAvatar recovers fine-grained facial and hand details.}
    \label{fig:qual_refinement}
\end{minipage}
\hfill
\begin{minipage}[t]{0.51\linewidth}
    \centering
    \includegraphics[width=\linewidth]{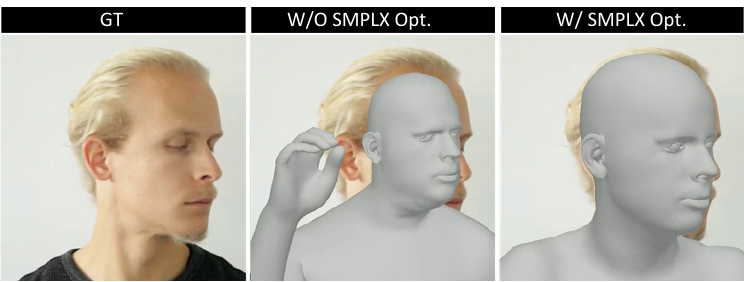}
    \caption{Effect of occlusion-robust SMPL-X optimization. Our 
confidence-weighted joint masking recovers a well-aligned 
identity-specific template.}
    \label{fig:qual_smplx}
\end{minipage}
\end{figure}

\noindent\textbf{w/o SMPL-X optimization.} Removing confidence-weighted 
keypoint masking leads to the largest single-component drop as shown in \cref{tab:insta_ablation_avg}. Under head-only input, off-the-shelf estimators produce unreliable pose estimates for occluded joints; without visibility-masked refinement these errors corrupt the canonical template and misaligned Gaussian attachment points, degrading even fully observed facial regions. The qualitative impact is shown in~\cref{fig:qual_smplx}.

\noindent
\textbf{w/o refinement.}
Removing the refinement heads substantially degrades reconstruction quality, particularly in high-frequency regions. As shown in~\cref{tab:neuman_ablation}, PSNR drops markedly, and the declines in SSIM and LPIPS further indicates the model’s reduced ability to synthesize sharp textures and fine geometric detail~\cref{fig:qual_refinement}.

\noindent
\textbf{w/o generated data.}
Excluding generated training views leads to artifacts when we render it from novel views, as shown in~\cref{fig:canonical_abalation}. This suggests that the generated data plays an important role in enhancing realism, improving texture completion for invisible region. We down-weight the diffusion-generated views' reconstruction loss to keep synthetic artifacts from overriding observed evidence (Supp. Sec. E.1, Tab. A2). To confirm these views preserve identity, we measure Face Consistency (FC), the ArcFace~\cite{ArcFace:CVPR2019} cosine similarity between rendered and ground-truth faces (Supp. Sec. E.2, Tab.A3), where our method achieves the highest FC among all baselines.

\vspace{-0.3cm}
\section{Conclusion}
\vspace{-2mm}

We present FlexiAvatar, a visibility-aware framework for reconstructing generalizable 3D Gaussian human avatars from monocular videos. In contrast to previous methods that implicitly assume full-body visibility, our approach restricts optimization to only the visible regions, effectively eliminating ghost artifacts and enhancing fidelity under partial-view inputs. Through the integration of SMPL-X-based body tracking, part-specific high-frequency refinement, visibility-aware optimization, and a diffusion-based generative completion module, FlexiAvatar provides a unified solution capable of high-quality reconstruction across full-body, upper-body, and head-only settings.

\noindent
\textbf{Limitations.}
Although FlexiAvatar delivers high-fidelity reconstructions across varying visibility conditions, it still shares a major limitation with existing 3DGS-based avatar methods~\cite{mononhrmonocularneuralhuman,Moon:ExAvatar:ECCV24,Qian:3DGSAvatar:CVPR24,Guo:Vid2Avatar:CVPR23}: modeling dynamically deforming clothing remains challenging. As an avenue for future work, our visibility-aware formulation could be extended to better identify and adaptively refine regions undergoing nonrigid deformations, enabling more accurate reconstruction of complex clothing dynamics.

\section*{Acknowledgements}
This work is supported by NRF grants (No. RS-2025-00521013 60\%, No. RS-2025-02216916 10\%) and IITP grants (No. RS2020-II201336 Artificial intelligence graduate school program(UNIST) 10\%; No. RS-2025-25442149 LG AI STAR Talent Development Program for Leading Large-Scale Generative AI Models in the Physical AI Domain 10\%), funded by the Korean government (MSIT). This work is also supported by the InnoCORE program of the Ministry of Science and ICT(26-InnoCORE-01) 10\%.
%
%
\bibliographystyle{splncs04}
\bibliography{main}
\clearpage
\appendix
\setcounter{section}{0}
\setcounter{figure}{0}
\setcounter{table}{0}
\setcounter{equation}{0}
\renewcommand{\tablename}{Tab.}

\title{FlexiAvatar: Unified 3D Gaussian Human Avatars\\ Under Arbitrary Body Visibility \\ \large Supplementary Materials} 

\titlerunning{FlexiAvatar: Unified 3D Gaussian Human Avatars}


\author{Yihalem~Yimolal~Tiruneh\inst{1} \and
Muhammad~Salman~Ali\inst{1} \and
Uyoung~Jeong\inst{1} \and
Muneeb~A.~Khan\inst{1} \and
MD~Khalequzzaman~Chowdhury~Sayem\inst{1} \and
Allanur~Bayramgeldiyev\inst{1} \and
Binod~Bhattarai\inst{2,3,4} \and
Seungryul~Baek\inst{1}}

\authorrunning{Y.~Tiruneh et al.}

\institute{%
{\small
$^{1}$UNIST, South Korea \enspace
$^{2}$University of Aberdeen, UK \enspace
$^{3}$University College London, UK \enspace
$^{4}$Fogsphere, UK}
\\[0.3em]
{\small\url{https://yihalem1.github.io/FlexiAvatar/}}
}
\setcounter{section}{0}
\renewcommand{\thesection}{\Alph{section}}

\setcounter{figure}{0}
\renewcommand{\thefigure}{A\arabic{figure}}

\setcounter{table}{0}
\renewcommand{\thetable}{A\arabic{table}}

\setcounter{equation}{0}
\renewcommand{\theequation}{A\arabic{equation}}

\maketitle

This supplementary material provides additional implementation details, quantitative evaluations, and visual comparisons complementing the main paper. It is organized as follows:
\begin{itemize}
    \item \textbf{Section~\ref{sec:smplx}}: Provides a detailed explanation of the SMPL-X registration procedure, including FLAME-based initialization and all associated loss components.
    \item \textbf{Section~\ref{sec:video_gen}}: Details the generative strategy for synthesizing auxiliary videos to handle unobserved textures.
    \item \textbf{Section~\ref{sec:network}}: Specifies the architectures of the Triplane-conditioned MLPs and the Part-Specific Residual MLP modules.
    \item \textbf{Section~\ref{sec:training}}: Enumerates the full set of training hyperparameters, including learning rates, optimization settings, and loss weight configurations.
    \item \textbf{Section~\ref{sec:ablation}}: Provides additional ablation studies evaluating the impact of synthetic-view loss weighting, Gaussian position refinement, and the spatial scope of the residual refinement module, the robustness of the Otsu visibility threshold, and a fine-grained face/hand reconstruction comparison against ExAvatar.
    \item \textbf{Section~\ref{sec:visuals}}: Presents additional qualitative results along with the accompanying video and extended figures.
\end{itemize}

\section{SMPL-X Registration Details}
\label{sec:smplx}

As outlined in Sec. 3.1 of the main paper, we perform a robust co-registration of SMPL-X parameters to obtain a personalized template that aligns accurately with the input video. 

\paragraph{FLAME Initialization and Registration.}
Since the standard SMPL-X shape space often lacks the fidelity required for high-quality facial animation, we leverage the FLAME~\cite{Li:FLAME:SA17} model as a strong prior. We first extract initial FLAME parameters (shape, expression, and jaw pose) using DECA~\cite{Feng:DECA:TOG21}. These parameters serve as an initialization for a subsequent fitting stage, where we optimize the FLAME mesh to align with 2D facial keypoints detected by an off-the-shelf regressor ~\cite{dwpose}. The optimization objective minimizes the $L_1$ distance between projected FLAME landmarks and detected 2D keypoints, regularized by an initialization loss to prevent deviation from the robust DECA prediction. 

\paragraph{SMPL-X Co-Registration.}
After completing facial registration, we perform full-body SMPL-X fitting. We initialize the body parameters using standard off-the-shelf estimators~\cite{smplest, dwpose}. To incorporate the high-fidelity facial detail recovered in the previous stage, we replace the expression parameters of the initial SMPL-X estimate with those obtained from the optimized FLAME model. Because FLAME and SMPL-X share a compatible expression basis, this substitution seamlessly transfers the accurate facial articulation predicted by DECA into the unified body model.

\begin{figure}[t!]
    \centering
    \includegraphics[width=0.9\linewidth]{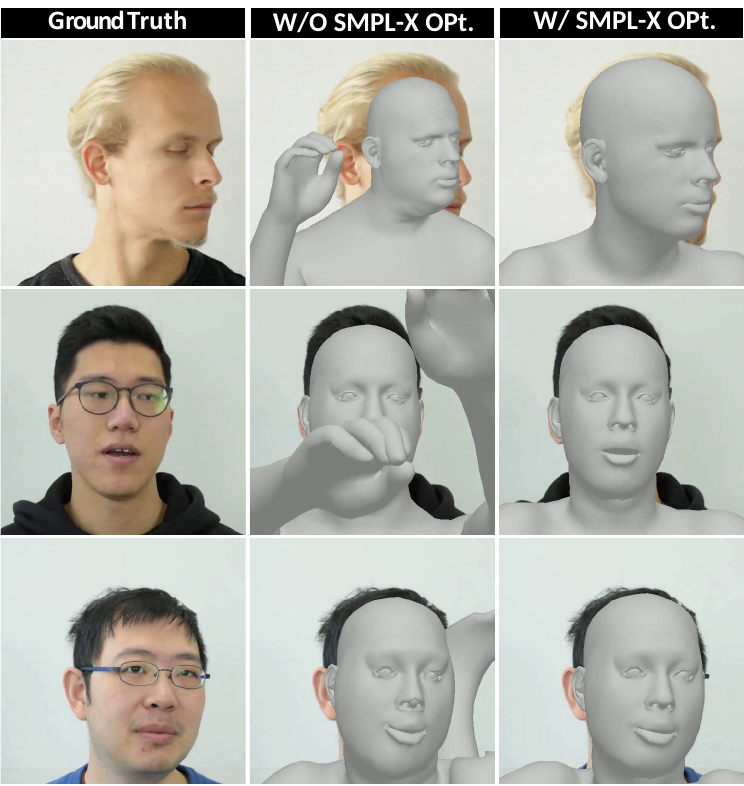}
    \caption{SMPL-X registration with and without optimization.}
    \label{fig:SMPL-X Registration Comparison}
\end{figure}

To obtain a personalized canonical template, we augment the SMPL-X representation with learnable offsets, joint offsets $\Delta J$, and facial vertex offsets $\Delta V_{face}$. The joint offsets modify the T-pose skeleton to better reflect subject-specific bone proportions, thereby influencing the rest-pose geometry. The facial offsets refine the surface mesh to match the FLAME geometry more closely. Qualitative results highlighting the improvements from our registration procedure are shown in~\cref{fig:SMPL-X Registration Comparison} and the supplemental video.

\paragraph{Hyperparameters and Loss Terms.}

The total registration objective $\mathcal{L}_{regi}$ (Eq. 1 in the main paper) is defined as:
\begin{equation}
\mathcal{L}_{regi}=\mathcal{L}_{kpt}+\lambda_{init}\mathcal{L}_{init}+\lambda_{sface}\mathcal{L}_{sface}+\lambda_{sreg}\mathcal{L}_{sreg}.
\end{equation}
where $\lambda_{init} = 0.1$, $\lambda_{sface} = 1.0$, and $\lambda_{reg} = 1.0$.  This objective balances keypoint alignment, initialization consistency, facial fitting, and structural regularization during registration.

\section{Video Generation for Unseen Texture}
\label{sec:video_gen}

As introduced in Sec. 3.2 of the main paper, our framework employs a generative strategy to synthesize auxiliary videos. This step is crucial for providing texture for unobserved regions in partial-view scenarios (e.g., upper-body or head-only clips). Our approach utilizes the diffusion-based human animation model MimicMotion~\cite{Zhang:MimicMotion:ICML25}, inspired by recent methods like PERSONA~\cite{sim2025persona}, to create training data from a single image. The core purpose is to compensate for limited pose information and reveal how appearance deforms across a wider range of motion. We use a target pose sequence along with the input image as input to the MimicMotion~\cite{Zhang:MimicMotion:ICML25} generator. 

We employ the diffusion-based generative strategy with dataset-specific adjustments. For each subject, we synthesize roughly 500 auxiliary video frames. In the full-body NeuMan dataset~\cite{Jiang:NeuMan:ECCV22}, we use a predefined pose trajectory that induces a full $360^{\circ}$ rotation, enabling comprehensive appearance coverage. In contrast, directly generating upper-body or head-only sequences for the TalkShow~\cite{Yi:TalkShow:CVPR23} and INSTA~\cite{INSTA:CVPR23} datasets leads to significant quality degradation. To address this challenge, we adopt a \textit{full-body-then-crop} strategy: we first generate a high-quality full-body sequence and subsequently crop the frames to match the visibility patterns of each dataset (upper-body for TalkShow and head-only for INSTA). This indirect approach leverages the stability of full-body generative models while still producing visibility-consistent texture priors for our optimization framework.

\section{Network Architecture Details}
\label{sec:network}

This section provides the specific implementation details for the Multi-Layer Perceptrons (MLPs) used in our framework, as introduced in Sections 3.3 and 3.5 of the main paper. 

\subsection{Triplane-Conditioned MLPs (Sec. 3.3)}

As outlined in Sec. 3.3, our decoder consists of two types of MLPs that operate on features extracted from the learnable triplane field $T$. All MLPs follow a shared backbone architecture comprising four fully connected layers with a hidden dimension of 128. Each hidden layer is followed by Group Normalization and a ReLU activation. The two MLP classes differ only in their respective input feature sets and output predictions.

\paragraph{Pose-Independent MLPs.}

This group of MLPs predicts the static, identity-dependent Gaussian parameters. Each MLP takes the interpolated triplane feature $f_i$ as input. The Static Geometry MLP outputs both the static vertex offset $\Delta V_{tri,i}$  (3D) and the static scale $ S_{\text{tri}, i}$ (3D), enabling subject-specific adjustments to the canonical geometry. The Static Color MLP produces the corresponding static color logit $C_{\text{tri}, i}$ (3D), providing a stable appearance prior for each Gaussian. Together, these MLPs encode the identity-specific geometry and color components independent of pose.

\paragraph{Pose-Dependent MLPs.}
This class of MLPs models dynamic, pose-dependent offsets conditioned on the articulated body configuration. The Pose-Dependent Geometry MLP takes as input the concatenation of the interpolated triplane feature $f_i$ and the SMPL-X body pose parameters $\theta$ (excluding the root). It predicts the pose-induced geometric adjustments, namely the dynamic vertex offset $\Delta V_{\text{pose}, i}$ (3D) and dynamic scale offset $\Delta S_{\text{pose}, i}$ (3D). Complementarily, the Pose-Dependent Color MLP receives the triplane feature $f_i$, the full set of body pose parameters $\theta$, and the local vertex normal $n_i$, and outputs the dynamic color offset $\Delta C_{\text{pose}, i}$ (3D). These MLPs together capture fine-grained, articulation-dependent variations in both geometry and appearance.

\subsection{Part-Specific Residual MLPs (Sec. 3.5)}

As detailed in Sec. 3.5, we introduce three lightweight MLPs ($F_{\theta}$) that specialize in predicting high-frequency color residuals for the face, left hand, and right hand. Designed to act as compact refinement modules, these MLPs are intentionally smaller than the triplane-conditioned networks. Each residual MLP comprises four fully connected layers with a hidden dimension of 64, and employs Group Normalization and ReLU activations. Following Eq. 7 in the main paper, the input to each network is the concatenation of the positional encoding $\gamma(\mu)$ of the Gaussian center, the positional encoding $\gamma(t)$ of the frame index, and the body pose $\theta$. The network outputs a 3D color residual $\Delta C_{\text{res}}$ tailored to the corresponding body region, enabling fine-grained, high-frequency appearance refinement.

\section{Training Hyperparameters}
\label{sec:training}

Our method is implemented in PyTorch~\cite{Paszke:PyTorch:NIPS19} and trained using the Adam optimizer~\cite{Kingma:Adam:ICLR15}. We use an initial learning rate of $1 \times 10^{-3}$ for all neural components, including the triplanes and MLPs, while the learnable SMPL-X parameters are optimized with a reduced learning rate of $1 \times 10^{-4}$. The overall training objective $\mathcal{L}$ (Eq. 9 in the main paper) combines reconstruction and regularization losses. The reconstruction loss employs weights of $\lambda_{L1} = 0.8$,  $\lambda_{ssim} = 0.2$, and $\lambda_{lpips} = 0.2$, while the facial consistency term is weighted by $\lambda_{face} = 0.8$. All regularization losses $\mathcal{L}{reg}$ use a weight of $\lambda_{reg} = 1.0$. As noted in Sec. 3.6, we differentiate between captured and generated data during training: captured frames use the full supervision weights, whereas the auxiliary generated frames employ down-weighted color losses ($\lambda_{L1}$ = $\lambda_{ssim}$ = $\lambda_{lpips} = 0.1$) to mitigate generative artifacts.

Training begins with a 2,000-iteration warm-up stage, during which only the coarse triplane-conditioned Gaussian representation (Sec. 3.3) is optimized to establish a stable initialization. After the warm-up, we activate two refinement components: (1) the part-specific residual MLPs $F_{\theta}$ (Sec. 3.5) to capture high-frequency details, and (2) the visibility-aware optimization module (Sec. 3.4), which filters Gaussians with weak or unreliable evidence based on accumulated visibility statistics.

Quantitative comparisons for head-only avatar reconstruction on the INSTA dataset are reported in ~\cref{tab:insta_exavatar_ours}, where our method consistently surpasses ExAvatar~\cite{Moon:ExAvatar:ECCV24}.
\begin{table}[ht]
\centering
\small
\caption{
Quantitative comparison on the INSTA (Head-only) dataset. We report the average performance over 4 subjects (Bala, Marcel, NF\_01, NF\_03). 
}
\label{tab:insta_exavatar_ours}
\begin{tabular}{lccc}
\toprule
Method & PSNR$\uparrow$ & SSIM$\uparrow$ & LPIPS ($\times 100$)$\downarrow$ \\
\midrule
ExAvatar~\cite{Moon:ExAvatar:ECCV24} & 29.07 & 0.924 & 9.66 \\
\textbf{Ours}                        & \textbf{32.11} & \textbf{0.946} & \textbf{6.97} \\
\bottomrule
\end{tabular}
\end{table}

\section{Additional Ablation Studies}
\label{sec:ablation}
To further validate the architectural and optimization choices within the FlexiAvatar framework, we present additional ablation studies focusing on loss weighting, the design of the residual refinement module and the robustness of using Otsu's visibility threshold. 

\subsection{Impact of Synthetic-View Loss Weight}
As discussed in Section 3.6 of the main paper, we additionally augment the training data with diffusion-generated views to provide texture cues for unobserved regions. We evaluate the sensitivity of the model to the weighting of the reconstruction loss applied to these synthetic frames. As shown in \cref{tab:synthetic_weight}, treating synthetic and captured views equally (weight = 1.0) degrades the overall reconstruction quality, leading to a PSNR drop to 33.59 dB. This degradation occurs because diffusion-based novel view synthesis can occasionally introduce minor structural artifacts or temporal inconsistencies. Down-weighting the synthetic-view loss to 0.1 provides sufficient gradient signal to regularize entirely unseen regions (e.g., the back of the avatar) without overriding the high-fidelity appearance constraints derived from the ground-truth captured frames. 

\begin{table}[ht]
\centering
\small
\caption{Impact of synthetic-view loss weight. Evaluated on the NeuMan dataset.}
\label{tab:synthetic_weight}
\begin{tabular}{lccc}
\toprule
Method & PSNR$\uparrow$ & SSIM$\uparrow$ & LPIPS ($\times 100$)$\downarrow$ \\
\midrule
Weight = 1.0        & 33.59 & 0.980 & 1.57 \\
Weight = 0.1 (\textbf{Ours}) & \textbf{35.77} & \textbf{0.987} & \textbf{0.83} \\
\bottomrule
\end{tabular}
\end{table}

\subsection{Identity Preservation of Generated Views}
A natural concern with the diffusion-generated auxiliary views (Sec.~3.2) is whether they preserve the subject's true identity or instead introduce a plausible but incorrect appearance. To evaluate this, we measure Face Consistency (FC), defined as the cosine similarity between ArcFace~\cite{ArcFace:CVPR2019} identity embeddings extracted from the rendered and ground-truth faces. For each method we render the reconstructed avatar at the test poses, detect and crop the face, extract a 512-dimensional ArcFace embedding, and average the cosine similarity to the corresponding ground-truth embedding over all test frames; higher FC indicates stronger identity preservation. As reported in~\cref{tab:fc_comparison}, our method attains the highest FC (0.7903) among all baselines on NeuMan~\cite{Jiang:NeuMan:ECCV22} while simultaneously achieving the best PSNR. This confirms that our generated views act as appearance regularization for unobserved regions without degrading subject identity in the observed regions.

\begin{table}[ht]
\centering
\small
\setlength{\tabcolsep}{6pt}
\caption{Face Consistency (FC) comparison on the NeuMan~\cite{Jiang:NeuMan:ECCV22} dataset.}
\label{tab:fc_comparison}
\begin{tabular}{lcc}
\toprule
Method & FC$\uparrow$ & PSNR$\uparrow$ \\
\midrule
Vid2Avatar     & 0.7221 & 30.70 \\
Vid2AvatarPro  & 0.7726 & 32.71 \\
ExAvatar       & 0.7631 & 34.80 \\
\textbf{Ours}  & \textbf{0.7903} & \textbf{35.77} \\
\bottomrule
\end{tabular}
\end{table}
\subsection{Impact of Gaussian Position Refinement}
In Sec. 3.5 of the main paper, we introduce part-specific residual MLPs that are designed solely to refine high-frequency appearance (color) details. Importantly, this module does not modify the Gaussian mean positions. To justify this design choice, we perform an ablation in which the residual refinement network is additionally allowed to predict unconstrained 3D positional offsets together with color residuals.

As shown in \cref{tab:position_refinement}, allowing the residual module to refine Gaussian mean positions leads to a decline in both perceptual quality and PSNR. In our hybrid representation, the Gaussians are explicitly anchored to the tracked SMPL-X surface topology, where geometric deformations are already modeled through the triplane representation and pose-conditioned offsets. Consequently, limiting the residual heads to appearance refinement enables the recovery of fine-grained texture details while preserving the structural stability of the underlying geometry.
 
\begin{table}[ht]
\centering
\small
\caption{Impact of Gaussian mean position refinement. Evaluated on the NeuMan dataset.}
\label{tab:position_refinement}
\begin{tabular}{lccc}
\toprule
Method & PSNR$\uparrow$ & SSIM$\uparrow$ & LPIPS ($\times 100$)$\downarrow$ \\
\midrule
\textbf{Ours}            & \textbf{35.77} & \textbf{0.987} & \textbf{0.83} \\
+ mean pos. refine       & 35.14 & 0.986 & 0.92 \\
\bottomrule
\end{tabular}
\end{table}

\subsection{Impact of Whole-Body Residual Refinement}
We further analyze the effect of extending the spatial coverage of the residual refinement module. In our full model, the residual MLPs are applied only to the face and hands to capture complex, high-frequency details such as wrinkles and subtle expressions. As an alternative, we evaluate a variant in which residual refinement is applied to the entire body by introducing an additional residual MLP that also refines the torso and limb regions.
As reported in \cref{tab:body_refine}, extending residual refinement to the full body provides negligible improvement in PSNR (35.78 vs. 35.77) while slightly worsening the perceptual LPIPS score (0.84 vs. 0.83). This observation indicates that the base triplane representation and pose-conditioned deformation networks already have sufficient capacity to model the geometry and appearance of the torso and limbs. Incorporating an additional residual module for the entire body therefore increases optimization complexity and computational overhead without yielding meaningful visual benefits, supporting our design choice of targeted, part-specific refinement.

\begin{table}[ht]
\centering
\caption{Impact of whole-body residual refinement. Average image metrics (Citron, Seattle, Jogging, Bike) evaluated on the NeuMan dataset.}
\label{tab:body_refine}
\begin{tabular}{lccc}
\toprule
Method & PSNR$\uparrow$ & SSIM$\uparrow$ & LPIPS ($\times 100$)$\downarrow$ \\
\midrule
\textbf{Ours}            & 35.77 & \textbf{0.987} & \textbf{0.83} \\
+ body refine            & \textbf{35.78} & \textbf{0.987} & 0.84 \\
\bottomrule
\end{tabular}
\end{table}

\subsection{Robustness of the Otsu Visibility Threshold
}

Our visibility-aware optimization (Sec. 3.4) uses Otsu's method \cite{otsu1979threshold} to automatically determine the threshold $\tau^*$ that separates visible from invisible Gaussians, avoiding the need for manual tuning. To validate this design choice, we examine the empirical distribution of per-Gaussian observation scores and compare Otsu against fixed alternatives.

\begin{figure}[t]
    \centering
    \includegraphics[width=0.7\linewidth]{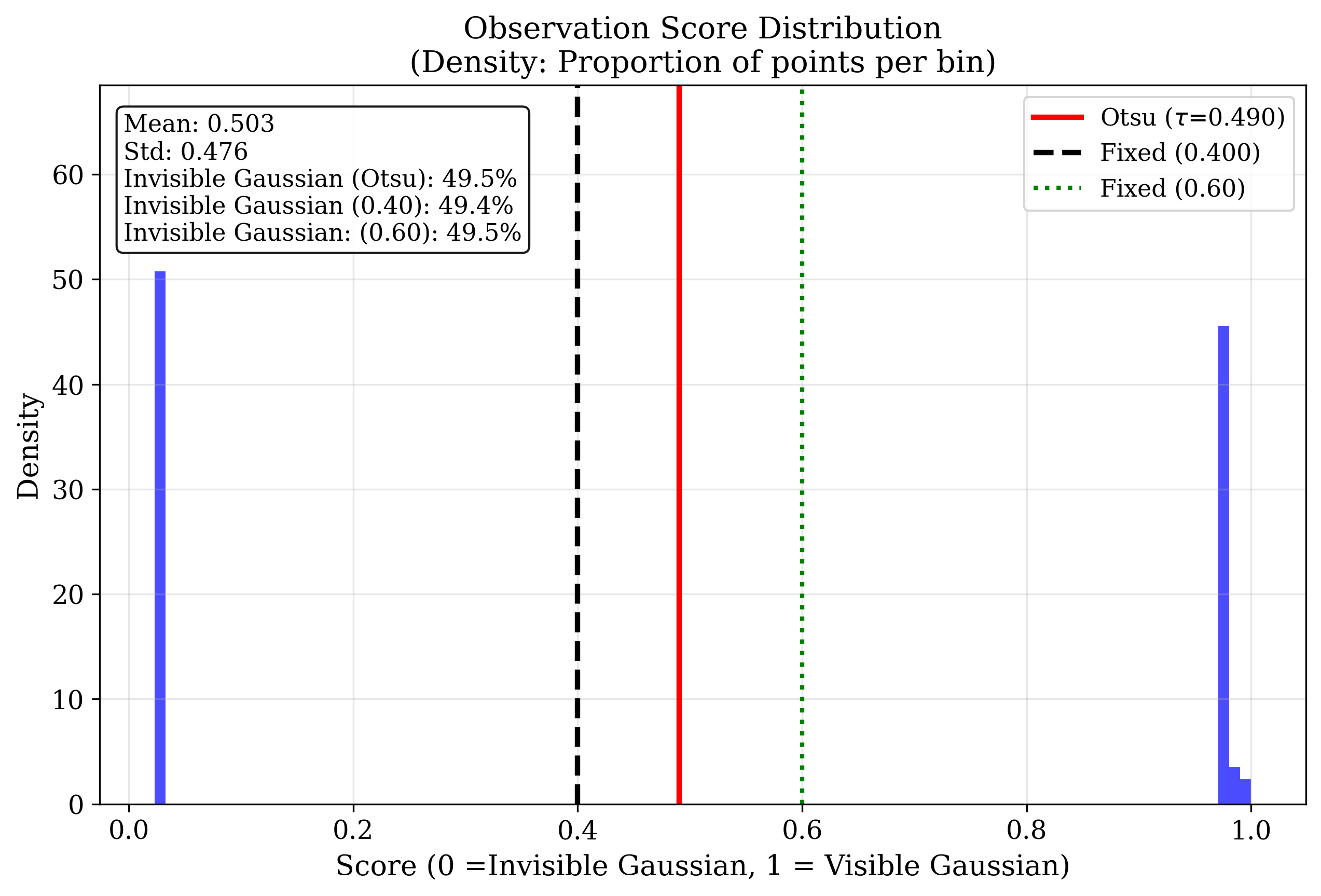}
    \caption{Empirical distribution of per-Gaussian observation scores, showing the bimodal structure exploited by Otsu's method.}
    \label{fig:otsu}
\end{figure}

As shown in ~\cref{fig:otsu}, the score distribution is strongly bimodal: approximately 49\% of Gaussians cluster near 0 (never observed) and the remainder cluster near 1 (consistently observed), with virtually no mass in the intermediate range. This bimodality is precisely the condition under which Otsu's method is optimal; it minimizes intra-class variance and reliably recovers the natural separation between the two populations. Because the two modes are so well-separated, the resulting threshold ($\tau^*$=0.490) is largely insensitive to small perturbations, and fixed alternatives such as 0.4 or 0.6 enclose essentially the same set of invisible Gaussians (49.4\% and 49.5\%, respectively).
~\cref{tab:otsu_vs_hardthreshold_biden} confirms this numerically on the head-only Biden sequence from INSTA\cite{INSTA:CVPR23} dataset. The automatically derived Otsu threshold achieves performance virtually identical to a carefully manually-tuned fixed threshold ($0.400$). The practical advantage of Otsu's method is therefore its adaptivity: it requires no dataset-specific tuning and automatically adapts to sequences with different proportions of visible body coverage, making it the preferred choice for a unified pipeline that spans full-body, upper-body, and head-only inputs.

\begin{table}[t]
\centering
\small
\setlength{\tabcolsep}{6pt}
\caption{Effect of fixed vs. Otsu threshold on Head-only (Biden).}
\label{tab:otsu_vs_hardthreshold_biden}
\begin{tabular}{lccc}
\hline
Method & PSNR$\uparrow$ & SSIM$\uparrow$ & LPIPS ($\times 100$)$\downarrow$  \\
\hline
Ours (fixed Threshold (0.400)) & \textbf{33.52} &  \textbf{0.958} &  \textbf{4.18} \\
Ours (Otsu Threshold (0.490)) & 33.50 & \textbf{0.958} & \textbf{4.18} \\
\hline
\end{tabular}
\end{table}

\subsection{Fine-Grained Face and Hand Reconstruction}
    To quantify the benefit of our part-specific residual refinement (Sec.~3.5) in the most challenging high-frequency regions, we evaluate reconstruction quality on the cropped face and hand regions of the NeuMan~\cite{Jiang:NeuMan:ECCV22} dataset and compare against ExAvatar~\cite{Moon:ExAvatar:ECCV24}. As reported in~\cref{tab:neuman_face_hand}, our method consistently outperforms ExAvatar on both regions across all metrics, improving face PSNR from 21.71 to 22.27 dB and hand PSNR from 19.14 to 20.05 dB, with corresponding gains in SSIM and LPIPS. These improvements are consistent with the qualitative comparison in Fig.~7 of the main paper, confirming that the residual modules recover fine-grained appearance such as subtle facial expressions and finger creases that the global Gaussian representation alone cannot capture.

\begin{table}[ht]
\centering
\small
\setlength{\tabcolsep}{6pt}
\caption{Quantitative comparison of face and hand reconstruction on the NeuMan~\cite{Jiang:NeuMan:ECCV22} dataset.}
\label{tab:neuman_face_hand}
\begin{tabular}{llccc}
\toprule
Scope & Method & PSNR$\uparrow$ & SSIM$\uparrow$ & LPIPS ($\times 100$)$\downarrow$ \\
\midrule
\multirow{2}{*}{Face}
 & ExAvatar~\cite{Moon:ExAvatar:ECCV24} & 21.71 & 0.703 & 9.74 \\
 & \textbf{Ours} & \textbf{22.27} & \textbf{0.719} & \textbf{8.15} \\
\midrule
\multirow{2}{*}{Hands}
 & ExAvatar~\cite{Moon:ExAvatar:ECCV24} & 19.14 & 0.571 & 15.03 \\
 & \textbf{Ours} & \textbf{20.05} & \textbf{0.605} & \textbf{14.64} \\
\bottomrule
\end{tabular}
\end{table}

\section{Additional Experimental Results}
\label{sec:visuals}

This section provides extended quantitative evaluations and qualitative visual comparisons across all body-visibility settings, complementing the main paper.

\subsection{Per-Subject Quantitative Breakdowns}
For completeness, we report the full per-subject breakdowns underlying the averaged results presented in the main paper. Table~\ref{tab:wildavatar_persubject} reports per-subject results on the WildAvatar~\cite{wildavatar:CVPR2025} dataset against GauHuman~\cite{gauhuman:CVPR2024} and ToMiE~\cite{tomie:ICCV2025}. On both benchmarks, our method achieves the best average performance across all metrics.

\begin{table*}[t]
\centering
\setlength{\tabcolsep}{6pt}
\renewcommand{\arraystretch}{1.15}
\caption{Per-subject quantitative comparison on the WildAvatar~\cite{wildavatar:CVPR2025} dataset (7 randomly selected subjects). We report PSNR\,$\uparrow$, SSIM\,$\uparrow$, and LPIPS\,$\downarrow$. Best per column is shown in \textbf{bold}.}
\label{tab:wildavatar_persubject}
\resizebox{\linewidth}{!}{%
\begin{tabular}{lcccccccc}
\toprule
\textbf{Method} & \textbf{Subject 1} & \textbf{Subject 2} & \textbf{Subject 3} & \textbf{Subject 4} & \textbf{Subject 5} & \textbf{Subject 6} & \textbf{Subject 7} & \textbf{Avg} \\
\midrule
\multicolumn{9}{l}{\textbf{PSNR}\,$\uparrow$} \\
\midrule
GauHuman~\cite{gauhuman:CVPR2024} & 22.80 & 26.37 & 28.38 & 30.08 & 35.79 & 27.53 & 27.19 & 28.31 \\
ToMiE~\cite{tomie:ICCV2025}       & 22.61 & 28.62 & 28.31 & 31.00 & 35.74 & 27.11 & 27.19 & 28.65 \\
\textbf{Ours} & \textbf{23.09} & \textbf{33.36} & \textbf{31.20} & \textbf{35.67} & \textbf{35.84} & \textbf{28.77} & \textbf{28.92} & \textbf{30.98} \\
\midrule
\multicolumn{9}{l}{\textbf{SSIM}\,$\uparrow$} \\
\midrule
GauHuman~\cite{gauhuman:CVPR2024} & 0.883 & 0.966 & 0.965 & 0.978 & 0.980 & 0.959 & 0.965 & 0.957 \\
ToMiE~\cite{tomie:ICCV2025}          & 0.879 & 0.973 & 0.965 & 0.979 & 0.980 & 0.958 & 0.965 & 0.957 \\
\textbf{Ours} & \textbf{0.921} & \textbf{0.988} & \textbf{0.980} & \textbf{0.988} & \textbf{0.983} & \textbf{0.975} & \textbf{0.980} & \textbf{0.973} \\
\midrule
\multicolumn{9}{l}{\textbf{LPIPS}\,$\downarrow$} \\
\midrule
GauHuman~\cite{gauhuman:CVPR2024} & 10.96 & 3.02 & 3.16 & 1.63 & 1.34 & 3.83 & 2.89 & 3.83 \\
ToMiE~\cite{tomie:ICCV2025}          & 11.35 & 2.56 & 3.09 & 1.70 & 1.30 & 3.92 & 2.78 & 3.81 \\
\textbf{Ours} & \textbf{9.99} & \textbf{1.69} & \textbf{1.82} & \textbf{1.00} & \textbf{1.20} & \textbf{2.82} & \textbf{1.99} & \textbf{2.93} \\
\bottomrule
\end{tabular}%
}
\end{table*}

\begin{figure}[t!]
\centering
\includegraphics[width=\linewidth]{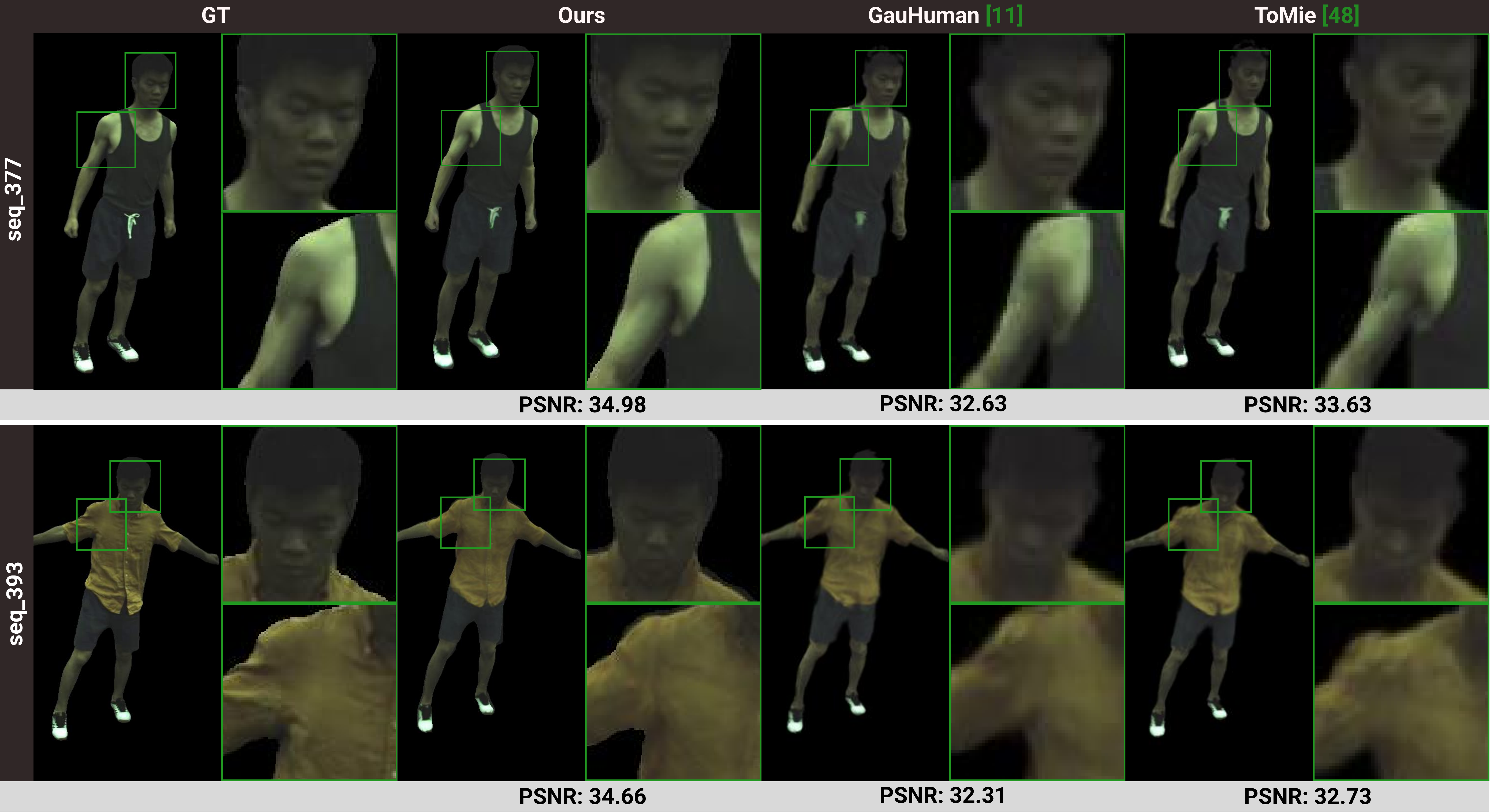}
\caption{Qualitative comparison on the ZJU-MoCap full-body dataset.}
\label{fig:supp_zju}
\end{figure}

\begin{figure}[t!]
\centering
\includegraphics[width=\linewidth]{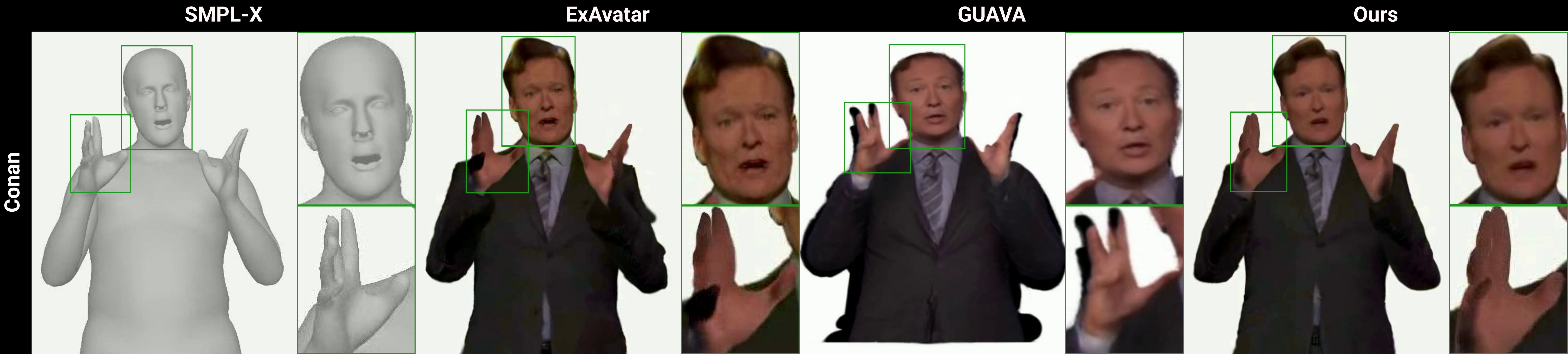}
\caption{Qualitative comparison on the TalkShow upper-body dataset.}
\label{fig:supp_talkshow}
\end{figure}

\begin{figure}[t!]
\centering
\includegraphics[width=\linewidth]{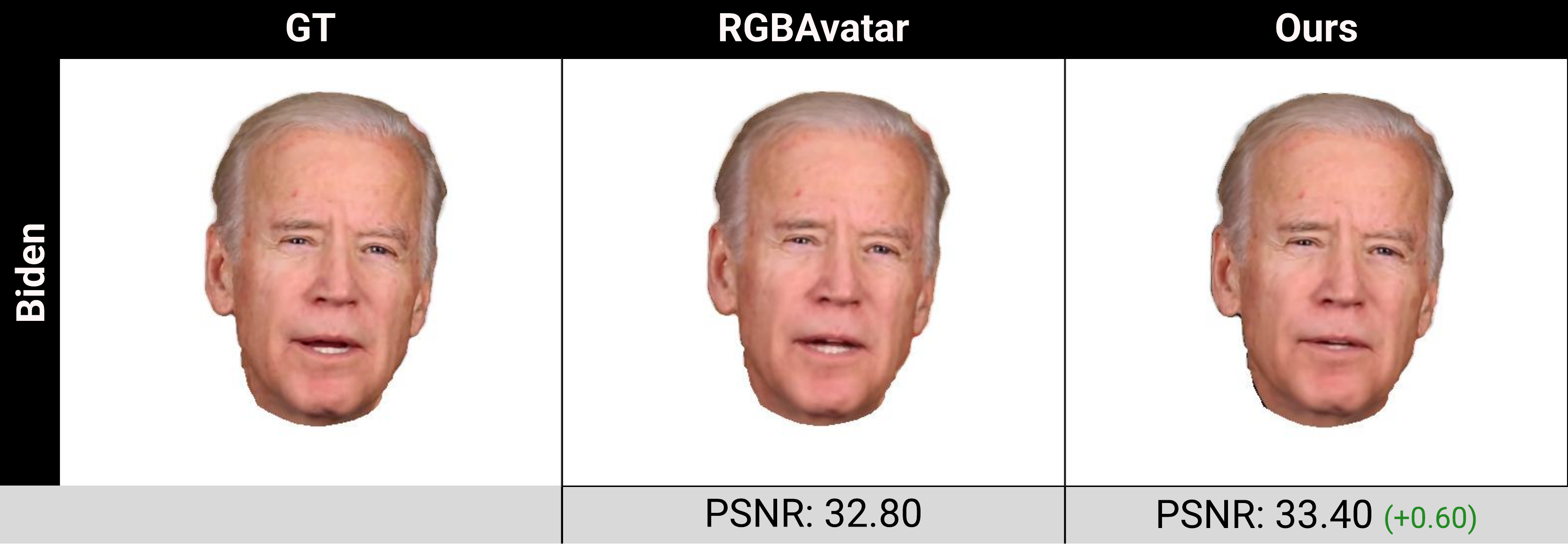}
\caption{Qualitative comparison on the INSTA head-only dataset.}
\label{fig:supp_insta}
\end{figure}

~\cref{fig:supp_zju} presents results on the ZJU-MoCap full-body dataset, where our method produces noticeably higher-quality reconstructions compared to prior methods, with improved texture fidelity and geometric accuracy.
~\cref{fig:supp_talkshow} shows upper-body comparisons on the TalkShow dataset against ExAvatar and GUAVA, highlighting clearer facial and hand details and more consistent appearance reconstruction.
~\cref{fig:supp_insta} provides qualitative results on the INSTA head-only dataset, demonstrating improved facial detail recovery compared to RGBAvatar.

\begin{figure}[t!]    \centering
    \includegraphics[width=\linewidth]{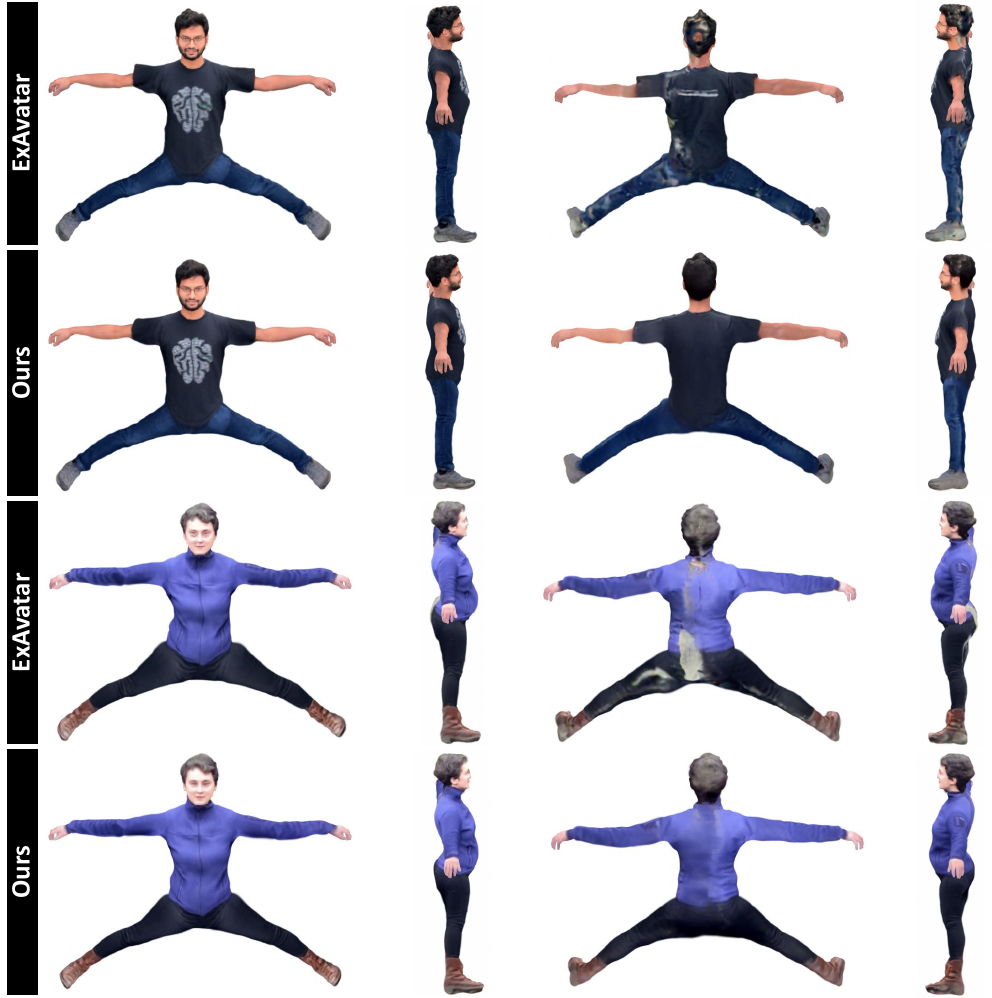}
    \caption{Qualitative comparison of our method and ExAvatar rendered in canonical space.}
    \label{fig:supp_cano_figure}
\end{figure}

\begin{figure}[t!]
    \centering
    \vspace{0.4cm}
    \includegraphics[width=0.8\linewidth]{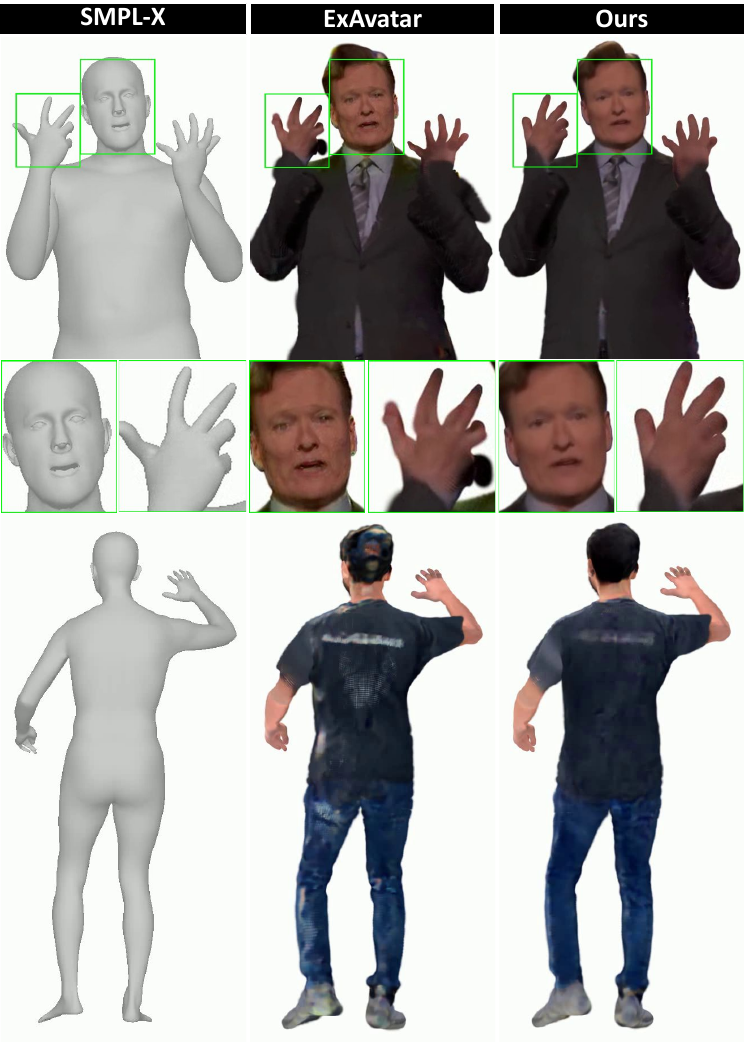}
    \caption{Side-by-side comparison of animated results under novel pose sequences produced by our method and ExAvatar.
    }
    \label{fig:supp_animation_figure}
    \vspace{-0.5cm}
\end{figure}

\begin{figure}[t!]
    \centering
    \vspace{0.4cm}\includegraphics[width=0.8\linewidth]{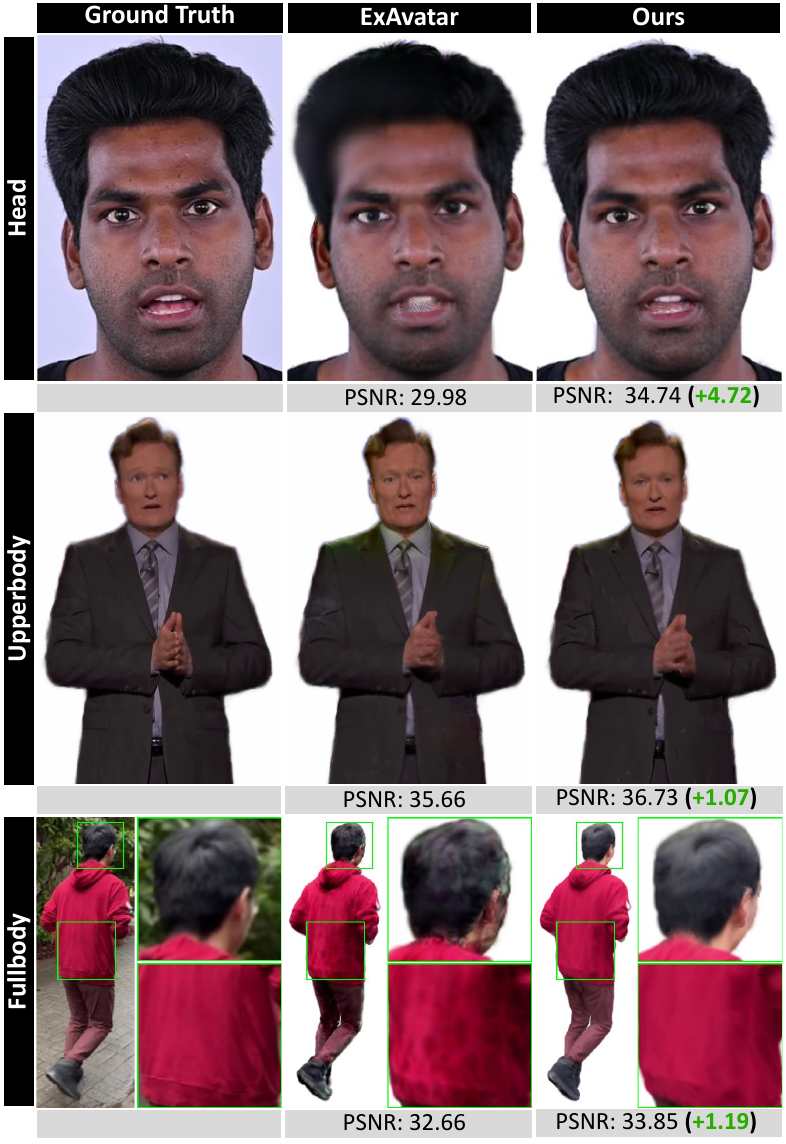}
    \caption{Qualitative comparison of Ours with ExAvatar under full-body, upper-body, and head-only inputs. Across all settings, our method yields noticeably sharper geometry, improved texture fidelity, and superior high-frequency detail reconstruction.}
    \label{fig:supp_qual}
    \vspace{-0.5cm}
\end{figure}

~\cref{fig:supp_cano_figure} visualizes the reconstructed 
canonical avatar in a T-pose with a full $360^{\circ}$ rotation, highlighting 
the completeness and global consistency of the recovered geometry and 
appearance. 
~\cref{fig:supp_animation_figure} provides novel-pose animations 
that demonstrate the model's ability to generalize to motion sequences not 
observed during training. 
Finally, ~\cref{fig:supp_qual} presents qualitative comparisons 
against baseline methods under full-body, upper-body, and head-only inputs, 
showcasing the superior texture fidelity and high-frequency detail preservation 
achieved by our approach.
For more qualitative comparisons, please refer to the supplementary video.


\end{document}